%% file: main.tex
\documentclass[sigconf,nonacm]{acmart}
\renewcommand\footnotetextcopyrightpermission[1]{}

\AtBeginDocument{%
  }

\input{preamble}

\begin{document}

\title{\framework: Semantic Feature-aware Robustness Testing of Deep Neural Networks}

% \affiliation{%
\author{Nusrat Jahan Mozumder}
\email{nm8tm@virginia.edu}
\orcid{0009-0003-1802-5150}
\affiliation{%
  \institution{University of Virginia}
  % \city{Charlottesville}
  % \state{Virginia}
  \country{USA}
}

\author{Divya Gopinath}
\email{divya.gopinath@nasa.gov}
\affiliation{%
  \institution{KBR Inc., NASA Ames}
  % \city{Mountain view}
  % \state{California}
  \country{USA}
}

\author{Corina Pasareanu}
\email{pcorina@cmu.edu}
\affiliation{%
  \institution{carnegie mellon university}
  % \city{Mountain view}
  % \state{California}
  \country{USA}
}

\author{Matthew Dwyer}
\email{matthewbdwyer@virginia.edu}
\affiliation{%
  \institution{University of Virginia}
  % \city{Charlottesville}
  % \state{Virginia}
  \country{USA}
}
% \affiliation{%

% \affiliation{%

% \affiliation{%

% \affiliation{%

% \affiliation{%

% \affiliation{%

% \affiliation{%

\renewcommand{\shortauthors}{mozumder et al.}
%%
%% Article type: Research, Review, Discussion, Invited or position
% \acmArticleType{Review}
%%
%% Links to code and data
% \acmCodeLink{https://github.com/borisveytsman/acmart}
% \acmDataLink{htps://zenodo.org/link}
%%
%% Authors' contribution
\acmContributions{BT and GKMT designed the study; LT, VB, and AP
  conducted the experiments, BR, HC, CP and JS analyzed the results,
  JPK developed analytical predictions, all authors participated in
  writing the manuscript.}
%%
%% Sometimes the addresses are too long to fit on the page.  In this
%% case uncomment the lines below and fill them accodingly.
%%
%% \authorsaddresses{Corresponding author: Ben Trovato,
%% \href{mailto:trovato@corporation.com}{trovato@corporation.com};
%% Institute for Clarity in Documentation, P.O. Box 1212, Dublin,
%% Ohio, USA, 43017-6221}

\begin{abstract}
\input{writing/abstract}
\end{abstract}


\settopmatter{printfolios=true}
\maketitle

% \hidecomments
%\input{writing/paper_plan}

\section{Introduction}
\input{writing/introduction}

\section{Related Work}
\input{writing/rel_work}

\section{Background}
\input{writing/background}

\section{Approach}
\label{sec:approach}
\input{writing/approach}

\section{Evaluation}
\label{sec:evaluation}

\input{writing/evaluation}
\section{Threats to validity.}
\input{writing/threats.tex}

\section{Conclusion}
\input{writing/Conclusion}

\section{Data Availability}
\input{writing/Data_availability}

% \section{Problem statement}
% In this document we discuss how to write an ACM article.4
% \section{Methods}

% This document provides \LaTeX\ templates for the article. We
% demonstrate different versions of ACM styles and show various options
% and commands.  We add extensive documentation for these commands and
% show examples of their use.

% \section{Results}

% We hope the resulting templates and documentation will help the
% readers to write submissions for ACM journals and proceedings.

% \section{Significance}

% This document is important for anybody wanting to comply with the
% requirements of ACM publishing.

\bibliographystyle{IEEEtran.bst}
\bibliography{main}
\end{document}

%% file: preamble.tex
\usepackage{amsfonts}
\usepackage{amsmath}
\usepackage{amsthm}
\usepackage{ulem}
\usepackage{cancel}
\usepackage{comment}
\usepackage{bm}
\usepackage{tikz}
\usepackage{algorithm}
\usepackage{algorithmicx}
\usepackage{algpseudocode}
\usepackage{graphicx}
\usetikzlibrary{calc}
\usetikzlibrary{intersections}
\usetikzlibrary{shapes}
\usetikzlibrary{shapes.multipart}
\usetikzlibrary{shapes.geometric}
\usetikzlibrary{arrows}
\usepackage{pgf}
\usepackage{pgfplots}
\pgfplotsset{compat=1.18}
\usepackage{pgfplotstable}
\usepgfplotslibrary{fillbetween}
\usepackage{tikz-3dplot}
\usetikzlibrary{3d}
\usepackage{wrapfig}
\usepackage{subcaption}
\usepackage{multirow}
\usepackage{subfiles}
\usepackage{lineno}
\usepackage{hyperref}
\usepackage{enumitem}
\usepackage{makecell}
\usepackage[vskip=1pt,leftmargin=2em]{quoting}
\usepackage{adjustbox}
\usepackage{listings}

\usepackage{xspace}
\usepackage{booktabs}
\usepackage{subcaption}
\usepackage[mode=buildnew]{standalone}

\usepackage[dvipsnames]{xcolor}

\algrenewcommand\algorithmicindent{0.5em}

\renewcommand{\rq}[1]{\textbf{RQ#1}}

\newcommand{\corina}[1]{({\color{orange} \bf Corina: #1})}
\newcommand{\nusrat}[1]{({\color{blue} \bf Nusrat: #1})}
\newcommand{\divya}[1]{{\color{red}\{Divya: #1\}}}
\newcommand{\ignore}[1]{}

\newcommand{\torepo}[1]{}

\usepackage{tikz}
\usepackage{pgfplots}
\usetikzlibrary{positioning,calc,arrows.meta,shapes,fit,backgrounds}
\usetikzlibrary{calc}
\usetikzlibrary{intersections}
\usetikzlibrary{shapes}
\usetikzlibrary{shapes.multipart}
\usetikzlibrary{shapes.geometric}
\usetikzlibrary{arrows}
\usepackage{tikz-3dplot}
\usetikzlibrary{3d}

\tikzset{
   operator/.style = {rectangle, thick, draw, rounded corners, minimum width=0.6cm, minimum height = 1cm},
   choice/.style = {diamond, thick, draw, inner sep=0},
   sgvertex/.style = {draw, circle, minimum width=10mm, thick}
}

\renewcommand{\sectionautorefname}{\S\kern-2pt}
\newcommand{\framework}{\textsc{SeFaR}\xspace}
\newcommand{\sfs}{FFT\xspace}

\usepackage{booktabs}
\usepackage{pifont}
\usepackage{tabularx}

\newcommand{\cmark}{\ding{51}}
\newcommand{\xmark}{\ding{55}}
\newcommand{\pmark}{$\circ$}

\usepackage{graphicx}
\usepackage{subcaption}

%% file: writing/abstract.tex
Deep neural networks are increasingly deployed in safety-critical domains as perception modules, 
where failures 
are often caused due to rare and under-represented scenarios. This necessitates the need to evaluate the \textit{semantic robustness} of perception models; conformance of behavior to high-level requirements over real-world perceptual variability. 
To address this, we propose \textbf{\framework}, a framework for systematic \textit{semantic-feature-centric testing} of vision models. Given a natural-language requirement and a set of satisfying inputs, \framework evaluates robustness with respect to diverse realistic semantic variations that preserve requirement satisfaction.
The approach employs a novel hierarchical concept model enabling structured exploration of the feature space and incorporation of domain knowledge via user-defined concepts. State-of-the-art diffusion and vision-language models are leveraged to generate photorealistic semantics-preserving perturbations and identification of \textit{previously unknown}  features impacting behavior. A feedback-driven adaptive process is adopted to generate interpretable \textit{failure-inducing semantic concepts along with corresponding test inputs}. Evaluation on case studies demonstrates that the proposed framework effectively satisfies requirement preconditions while identifying requirement-independent features that influence model decisions, enabling it to both uncover faults and relate them to such features. 

%% file: writing/introduction.tex
\begin{figure*}[t]
\centering

\begin{subfigure}{\linewidth}
\centering
\includegraphics[width=\linewidth]{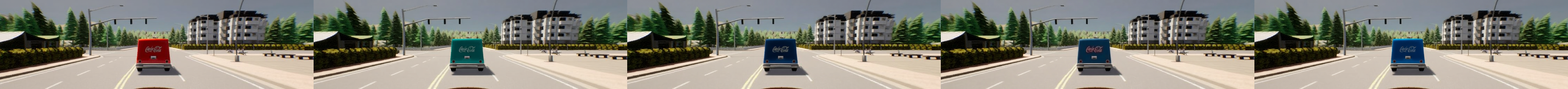}
\caption{vehicle color change}
\end{subfigure}

\ignore{
\begin{subfigure}{\linewidth}
\centering
\includegraphics[width=\linewidth]{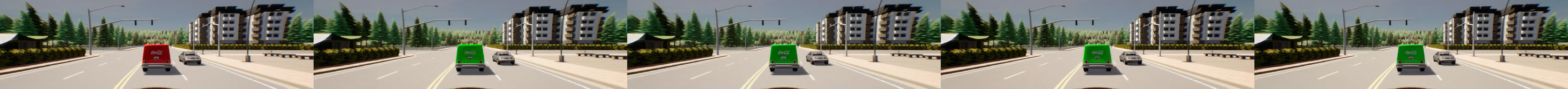}
\caption{vehicle color change to green}
\end{subfigure}
}

\begin{subfigure}{\linewidth}
\centering
\includegraphics[width=\linewidth]{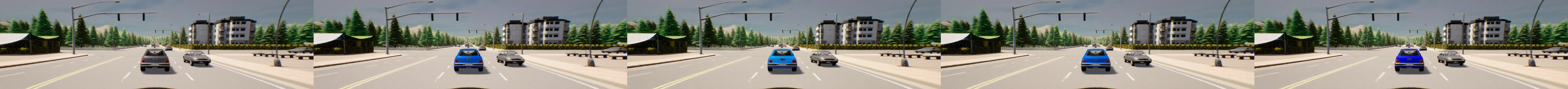}
\caption{vehicle color change to blue}
\end{subfigure}

\begin{subfigure}{\linewidth}
\centering
\includegraphics[width=\linewidth]{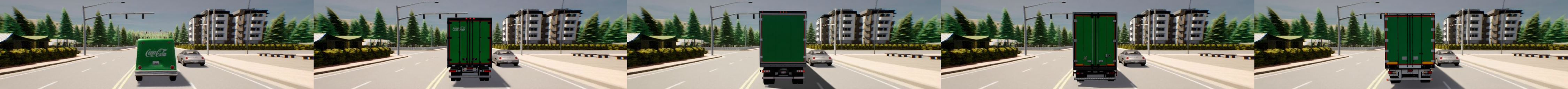}
\caption{vehicle type change to truck}
\end{subfigure}

\begin{subfigure}{\linewidth}
\centering
\includegraphics[width=0.7\linewidth]{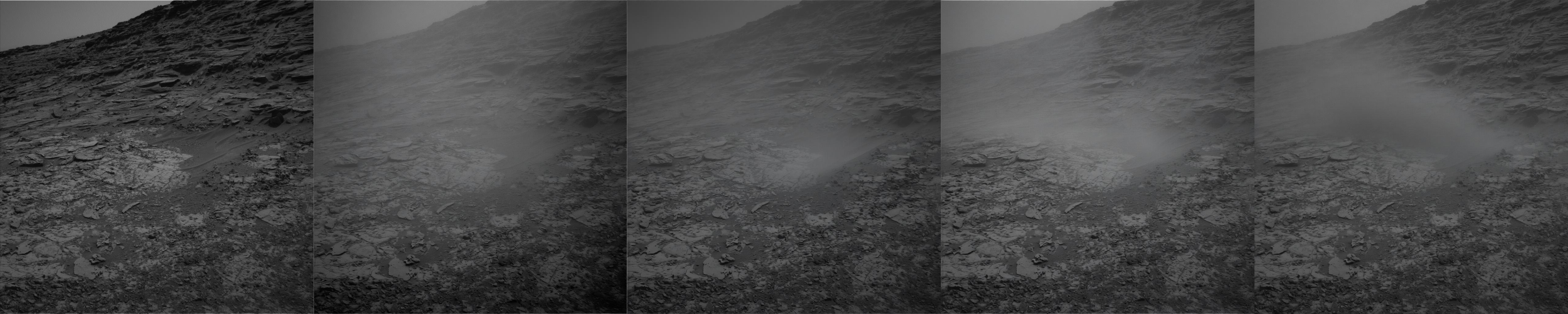}
\caption{weather condition change to dust-storm.}
\end{subfigure}
\vspace{-5mm}
\caption{Examples of \framework generated misclassified images and corresponding failure-inducing semantic features from SGSM  and AI4Mars case-studies. The leftmost image in each panel satisfies the respective requirement. The rest of the images show perturbations with respect to a specific concept while preserving other scene semantics, leading to incorrect model behavior.}
\label{fig:failure_cases}

\end{figure*}

Deep neural networks (DNNs) are widely deployed in safety-critical systems such as autonomous vehicles~\cite{zhang2023unsupervised,wu2022trajectoryguided,shao2023safety,toledo2021deeper,bojarski2016end,chen2020learning}, small unmanned aerial systems~\cite{giusti2016machine,loquercio2018dronet,pham2018deep}, and planetary exploration~\cite{proenca2019deep,van2018satellite,demir2018deepglobe,zhu2017deep,hundman2018detecting}. They aid in several safety-critical tasks such as perception-based collision avoidance, object detection and tracking, trajectory and motion control, terrain assessment and anomaly detection in uncertain environments. Despite strong empirical accuracy, DNNs have been shown to fail in rare, underrepresented, or semantically complex scenarios, leading to unsafe behavior~\cite{banks2017tesla,penmetsa2021crashes,abdelaty2024accident,silva2026crash}. It is indeed very challenging to validate performance across \textit{all} relevant scenarios, necessary for safety-critical systems~\cite{iso21448}, and this is even more acute in space applications where real-world data collection is extremely expensive and impractical~\cite{azzalini2023eventbased,zysk2023infrared}. This motivates the need for systematic testing approaches that support coverage of safety-relevant semantic conditions to help uncover failure-inducing scenarios.

\textit{Semantic features} are human-interpretable attributes of a scene that influence how it is perceived. They offer a natural abstraction for expressing system requirements and for reasoning about model behavior in human-interpretable terms. %Correspondingly, expected system behavior is often specified using requirements in terms of high-level, human-understandable semantic features. 
In safety-critical domains, it is essential to ensure that the system behavior meets specified requirements. However, this is challenging for systems with perception components that consume image data.

Prior research has proposed a range of testing and analysis techniques~\cite{dola2024cit4dnn,tian2018deeptest,pei2017deepxplore,guo2018dlfuzz,lee2020adaptive,wang2022bet,zohdinasab2022feature} to expose failures in DNN models for visual perception. However, many of these approaches focus on pixel-level transformations or variations that aim to exercise internal neuron activations or regions of latent representations, which may not correspond to meaningful changes.
%The behavior of perception models, however, often relies on meaningful semantic abstractions rather than low-level cues.
Approaches such as ~\cite{scenic-mlj23, verifai-cav19,HeRGT23, AlnaserSA21, FremontKPS20, woodlief2024s3c}, systematically explore structured scenario-level semantics using scene graphs and simulators, but are limited in capturing real-world perceptual variability. Hybrid methods combining simulators with generative models~\cite{AttaouiPB25, attaoui2025designator, baresi2025efficient} improve realism and diversity, however, are constrained by simulator parameterization, limiting their ability to explore unconstrained real-world semantics.

More recently, there has been a surge of techniques leveraging generative models to explore real-world semantic variations. The focus, however, has been on image editing~\cite{flux2024, fluxkontext2024, conceptsliders2024, qwenVL2023, yin2024pascalEA}, representation analysis~\cite{ConceptAlgebra2023, Conceptor2024, SeFa2021, NoiseCLR2024, Geodesics2025}, benchmark construction and augmentation~\cite{duenkel2025cns, Mofayezi_2023_CVPR, Zhang_2024_ImageNetD,dillema2025}, and post-hoc explainability\cite{prabhu2023lance, fathi2024decodex, conceptualedits2025}. They are not designed for systematic robustness evaluation of downstream vision models. The approach in \cite{mozumder2025rbt4dnn} evaluates vision model behavior with respect to requirements expressed in natural language. It leverages text-conditional diffusion to generate photorealistic images satisfying the precondition and checks for behavior conformance to the specified postcondition.
However, it does not support systematic assessment of model robustness with respect to unintended semantic variations irrelevant to the requirement. 
Perception models often rely on spurious correlations with unintended features. Evaluating robustness to such irrelevant semantic variations is crucial for uncovering brittle behaviors which can occur despite requirement satisfaction. The importance of this is amplified by recent evidences of failures in real-world safety-critical systems~\cite{NTSB2019UberCrash, NTSB2020TeslaCrash, NTSB2021TeslaTruck, NTSB2018TeslaFire}.\\

\noindent{\bf Our Approach.}
We propose \textbf{\framework}, a framework for systematic \textit{semantic-feature-centric testing} of vision models. Given a natural-language requirement describing desired model behavior of the form pre implies postcondition,  and a set of images that satisfy it, the framework evaluates the robustness of the model to perturbations along semantic features that are orthogonal to the preconditions.
%irrelevant \corina{maybe say orthogonal instead of irrelevant? they are orthogonal to preconditions but lead to failures so might violate postconditions, thus they are relevant?}\divya{added; please check} to requirement satisfaction; features that are orthogonal to the pre-condition and hence should not impact post-condition satisfaction. \corina{they should??} 
The key idea is to preserve semantics relevant to precondition satisfaction %\corina{change to precondition satisfaction} \divya{I think they are also relevant to the post-condition satisfaction. They are necessary but not sufficient conditions for postcond satisfaction.} 
(such as scene configuration, geometry, spatial relationships, and other task-relevant entities) while systematically generating realistic variations along other semantic dimensions, potentially uncovering rare, subtle, and previously unseen failure-inducing scenarios that may not be represented in available datasets.

\framework employs a novel \textit{hierarchical concept model} that organizes concepts in a top-down structure spanning across multiple levels of abstraction and granularity. This enables structured exploration ranging from variations in coarse factors (such as background, object type) to fine-grained attributes (such as color, pose).  Domain expertise is leveraged by allowing users to populate the hierarchy with relevant concepts, thereby enabling alignment of testing with knowledge of the operational domain. 
%It is typically easier for users to identify and specify general coarser features that are orthogonal to the pre-condition than finer attributes, and the hierarchical structure facilitates this. \corina{do we want to say something like: the hierarchy is populated by the user with coarse abstractions that are automatically refined?} \divya{addressed the first part. At this point, we have not yet mentioned about the automatic refinement or new feature discovery part.}

For each concept in the hierarchy, \framework leverages a state-of-the-art diffusion model~\cite{wu2025qwenimagetechnicalreport} to generate perturbed images that vary the selected feature while preserving rest of the scene semantics. The vision model is evaluated on the perturbed images and the misclassified and correctly classified inputs are analyzed using visual question answering~\cite{qwen_huggingface} to identify finer-grained semantic factors that correlate with misbehavior. These may correspond to user-specified concepts or to \textit{previously unknown semantic features} discovered through analysis.

This feedback is used for \textit{adaptive exploration} of the concept space and targeted test generation to progressively refine the characterization of failure-inducing conditions. The iterative process continues until the relevant depth of the hierarchy is covered, producing a structured set of \textit{failure-inducing semantic concepts} together with corresponding \textit{test inputs}. 
%\corina{same comment as in approach; as we discover new features we might not terminate so add something about a bound?}
%\divya{Nusrat, could you please address this?}\nusrat{I addressed you comment in the approach and added it, here I think the bound is covered by the relevand hierarchy. Please edit if you feel it unclear.ok}

\framework, thereby, presents a novel combination of hierarchical semantic modeling, diffusion-based generation, and feedback-guided exploration, to enable systematic discovery of scenarios which can cause safety-critical failures in real-world deployments. \\

\noindent\textbf{{Illustrative Examples.}} Let us consider the SGSM case-study~\cite{toledo2024sgsm}; a collection of real-world autonomous driving datasets with requirements derived from the Virginia Driving Code. Consider the requirement, \textit{\textbf{if a vehicle is within 10 meters, in front, and in the same lane, the ego will not accelerate}}. In Figure~\ref{fig:failure_cases}, the leftmost images in rows (a-c) are inputs where the system with the perception model satisfies the requirement, while the remaining are failure-inducing images generated by \framework. The user-defined root-level of the hierarchical model explored the replacement of the \textit{vehicle in front} of the ego. The images causing failure were analyzed for change with respect to the user-defined attributes \textit{vehicle color} and \textit{vehicle type}. The next iteration perturbed original images only for \textit{color}, and only for \textit{type}, respectively. Figure~\ref{fig:failure_cases}(a) shows example images leading to misbehavior where only the color of the vehicle was changed. Further exploration of 
of deeper portions of the \textit{color} feature tree revealed that 
specific colors, like the changes to \textit{blue} (Figure~\ref{fig:failure_cases}(b)), \textit{green}, and \textit{teal} consistently led to failure-inducing inputs. 
The color \textit{teal} was not in the original feature model and was
discovered by the analysis. 
Further exploration of the \textit{type} feature tree revealed that vehicle change to \textit{truck} emerged as the feature present in the most failure-inducing inputs (Figure~\ref{fig:failure_cases}(c)).

AI4MARS~\cite{swan2021ai4mars} is a large-scale dataset for terrain-aware autonomous navigation on Mars, collected from NASA’s rover missions. It consists of a large number of rover images annotated with pixel-level terrain labels. We trained a perception model that enables identification of big rocks in the scene and tested this model for the requirement, \textbf{\textit{if the MARS terrain contains big rocks then it should be detected.}} The image panel in Figure~\ref{fig:failure_cases}(d) highlights that a \textit{Dust Storm} is a failure-inducing weather condition; a novel scenario not present in the available data.\\
   % \begin{itemize}
   %     \item A semantic robustness testing framework that evaluates invariance to irrelevant semantic features.
   %     \item A systematic approach to identify features irrelevant to the requirement, but still influence the model output.
   %     \item An empirical evaluation on real-world safety-critical datasets demonstrating the effectiveness of the proposed approach.
   % \end{itemize}
   
\noindent\textbf{{Contributions.}} This work makes the following contributions:

\noindent\textbf{1. A  semantic-feature-centric testing framework}
, \framework, that systematically evaluates robustness of vision models with respect to semantic feature variation.

\noindent\textbf{2. A hierarchical concept model} that facilitates exploration of a semantic feature space, accommodates user-defined domain knowledge, and supports unspecified features.

\noindent\textbf{3. Diffusion-based generation of requirement-preserving semantic variations}
to detect  model non-robustness to variations in realistic, rare scenarios.

\noindent\textbf{4. Adaptive discovery of feature-oriented failure explanations}
through VQA-based semantic analysis of failure-inducing images and generation
of test sets that witness model feature non-robustness.

\noindent\textbf{5. Evidence of the applicability of \framework}. %and practicality 
%of the method in Section~\ref{sec:evaluation}.  %\matt{more here} 
Our experiments demonstrate that \framework is effective in generating tests that
satisfy requirement preconditions while identifying
new features that influence model decisions,
enabling it to both uncover faults and relate them to such features. 
%\divya{Nusrat, please check/add.}

%% file: writing/rel_work.tex
\begin{table*}[th]
\centering
\small
\setlength{\tabcolsep}{6pt}
\begin{tabularx}{\textwidth}{lcccccc}
\toprule

\textbf{Method Category} &
\makecell{\textbf{User}\\\textbf{Concepts}} &
\makecell{\textbf{Concept}\\\textbf{Model}} &
\makecell{\textbf{Diffusion}\\\textbf{Perturbations}} &
\makecell{\textbf{Used for}\\\textbf{Model Testing}} &
\makecell{\textbf{Failure}\\\textbf{Concepts Discovery}} &
\makecell{\textbf{Adaptive}\\\textbf{Exploration}} \\

\midrule

Robustness testing
{\cite{hu2024corruption, deeproad2018, ma2018deepgauge, deepxplore2017}} &
\xmark & \xmark & \xmark & \cmark & \xmark & \pmark \\
\midrule

Image editing 
{\cite{flux2024, fluxkontext2024, conceptsliders2024, qwenImage2025, yin2024pascalEA}} &
\cmark & \xmark & \cmark & \pmark & \xmark & \xmark \\
\midrule

Latent concept manipulation
{\cite{ConceptAlgebra2023, Conceptor2024, SeFa2021, NoiseCLR2024, Geodesics2025}} &
\pmark & \cmark & \pmark & \xmark & \xmark & \xmark \\
\midrule

Data augmentation 
{\cite{duenkel2025cns, Mofayezi_2023_CVPR, Zhang_2024_ImageNetD, dillema2025}} &
\pmark & \xmark & \cmark & \cmark & \xmark & \xmark \\
\midrule

Counterfactual explanations 
{\cite{prabhu2023lance, fathi2024decodex, conceptualedits2025}} &
\xmark & \pmark & \cmark & \pmark & \pmark & \xmark \\
\midrule

Requirement-guided testing 
{\cite{mozumder2025rbt4dnn}} &
\cmark & \xmark & \cmark & \cmark & \xmark & \pmark \\

\midrule
\textbf{\framework} &
\cmark & \cmark & \cmark & \cmark & \cmark & \cmark \\

\bottomrule
\end{tabularx}

\caption{Comparing approaches exploring real-world semantic variations impacting perception. Can \textit{User} provide \textit{Concepts}?, Is there a structure, \textit{Concept Model}, guiding concept-driven exploration?, Is a \textit{Diffusion} model used to generate \textit{Perturbations}?, Is the purpose to \textit{Test} the \textit{Model} on the perturbed images?, Are \textit{Failure} inducing \textit{Concepts} identified or \textit{Discovered}?, Is \textit{Adaptive Exploration} of concept space done to guide test generation. \pmark\ indicates partial support.}
\label{tab:concept_testing}

\end{table*}

There is considerable work in the realm of \textit{scenario-level} semantics and exploration of scenes with structured entities and relationships
(e.g., objects, positions, and interactions). Scenic~\cite{scenic-mlj23} is a probabilistic
 programming language for specifying scene configurations. Approaches such as ~\cite{verifai-cav19,HeRGT23, AlnaserSA21,woodlief2024s3c,FremontKPS20} 
 perform systematic scene generation, testing and falsification with respect to formal requirements in simulated environments. 
While these enable principled exploration and formal reasoning over the scenario space, they are typically constrained 
 by simulator fidelity and parameterization.
Failures in  deployment, however, are frequently induced by unexpected real-world
variations such as changes in real-world factors like weather, lighting, terrain, geomorphology and other environmental variations, which in turn impact perception model behavior. 
%It is therefore essential to evaluate the vulnerability of a model with respect to such perception-level semantics.
%Let $\mathcal{F}$ denote features in the visual attribute space. Given the input image space, $\mathcal{X}$, let $\mathcal{D}$ be a transformation (or perturbation) operator; 
%$\mathcal{D}: \mathcal{X} \times \mathcal{F} \rightarrow \mathcal{X}$, such that $\mathcal{D}(x,f)$ produces a 
%perturbed image where the visual attributes of $x$ are modified with respect to the feature, $f$. Let us consider a pre-condition, $\phi_x$ that captures the invariant scene semantics with respect to a post-condition $\phi_y$. A model $N$ is said to be robust at input $x \in \mathcal{X}$ with respect to $\mathcal{F}_x \subseteq \mathcal{F}$ (features present in $x$) if
%\begin{equation}
% \forall f \in \mathcal{F}_x,\quad 
%\phi_x(x) = \phi_x(\mathcal{D}(x,f)) 
%\;\Rightarrow\;
%\phi_y(N(x)) = \phi_y(N(\mathcal{D}(x,f))) 
%\label{eq:robustness_single}
%\end{equation}
%Let \(X_{\mathrm{rep}} \subseteq \mathcal{X}\) be a representative set of inputs. 
%A model \(N\) is said to be robust to variation \(f \in \mathcal{F}\) over \(X_{\mathrm{rep}}\) if
%\begin{equation}
%\forall x \in X_{\mathrm{rep}}, \;\; Eq~\ref{eq:robustness_single}\;\; holds. 
%\label{eq:robustness}
%\end{equation}
% Conversely, a model \(N\) is said to be vulnerable to variation \(f \in %\mathcal{F}\) over \(X_{\mathrm{rep}}\) if
%\begin{equation}
%\exists x \in X_{\mathrm{rep}} \;s.t.\;
%\phi_x(x) = \phi_x(\mathcal{D}(x,f))
%\;\wedge\;
%\phi_y(N(x)) \neq \phi_y(N(\mathcal{D}(x,f)))
%\label{eq:vulnerability}
%\end{equation}
Table~\ref{tab:concept_testing} presents approaches that explore such \textit{real-world level} semantics impacting perception. %\corina{what is perception-level semantics? perhaps you want to say "real-world semantics" as opposed to simulations?}

Traditional robustness testing of vision DNNs~\cite{goodfellow2015explaining, kurakin2017adversarial,papernot2017practical,hendrycks2019benchmarking,hu2024corruption, deeproad2018, ma2018deepgauge, deepxplore2017}
focus on low-level pixel perturbations, global corruptions (blur, brightness, noise) and affine transformations, with limited ability to reason about rich semantic variability. With the advent of generative models, recent approaches leverage them to generate photorealistic images. Diffusion specifically enables stable, high-quality generation of perturbations that are controllable via modalities such as text. Image-editing techniques~\cite{flux2024, fluxkontext2024, conceptsliders2024, qwenVL2023, yin2024pascalEA} leverage diffusion and vision-language models to perform fine-grained modifications of visual attributes. Approaches such as~\cite{ConceptAlgebra2023, Conceptor2024, SeFa2021, NoiseCLR2024, Geodesics2025} explore the latent space of generative models to identify
and manipulate directions to enable controlled variation of visual attributes. 
Benchmark generation and dataset augmentation approaches~\cite{duenkel2025cns, Mofayezi_2023_CVPR, Zhang_2024_ImageNetD,dillema2025} produce semantically diverse datasets 
covering variations in environmental conditions, style, and nuisance factors.  Counterfactual explanation generation \cite{prabhu2023lance, fathi2024decodex, conceptualedits2025} focuses on generating counterfactual images that alter
model predictions, providing localized insights into sensitivity to semantic features. A recent technique~\cite{mozumder2025rbt4dnn} systematically generates diverse test images that satisfy a given requirement in natural language. 

Table~\ref{tab:concept_testing} characterizes these approaches in terms of six factors.
The \textit{User Concepts} dimension evaluates if the respective method can leverage user knowledge, such as the concepts being provided by a domain expert or being elicited from requirements. More recent approaches accommodate this by leveraging text-conditional models to take in user input as prompts or natural language descriptions. 
%However, counterfactual explanation techniques typically perform data-driven discovery of concepts that lead to misprediction. Similarly, latent concept manipulation also discovers semantic directions from the model, while benchmark generation techniques operate along pre-defined semantic axes. 
A structured representation (or \textit{Concept Model}) aids in systematic exploration of the feature space. Concept manipulation along latent space directions does possess an implicit structure. However, most other approaches are not guided by structured concept-driven techniques.
Image editing and latent concept manipulation methods are primarily designed for computer vision tasks and representation analysis. They are not \textit{Used for Model Testing} to evaluate the robustness of downstream vision models. While others, such as benchmark generation, are indeed employed for testing vision models, they do not incorporate mechanisms for \textit{Failure Concept Discovery}. The approaches, \cite{prabhu2023lance, fathi2024decodex, conceptualedits2025}, do evaluate the model on perturbed images with the purpose of identifying minimal semantic variations or confounders that cause the decision to flip. However, they mostly serve as post-hoc explanations and do not perform \textit{Adaptive Exploration} of failure-inducing concepts.

\textbf{\framework} leverages a user-defined, potentially open, hierarchical concept model that enables specification and iterative exploration of concepts at varying granularities. The concept model can be populated by the user, as well as refined automatically by the discovery of new concepts.
It employs diffusion to generate photorealistic image variants that constitute model test inputs that witness non-robustness feature variation.
It performs failure analysis to identify features within the user-provided model that best explain failure and discovers novel features not included in the model. 
Finally, it uses the results of failure analysis to iteratively and adaptively explore finer-grain concepts to produce more accurate feature-based characterizations of failure.

%% file: writing/background.tex
%\subsection{Semantic Features-based Approaches}
\subsection{Multi-modal and Generative Models.}
%Generative models learn the underlying data distribution of natural images enabling them to produce photorealistic semantic variation. %Generative Adversarial %Networks (GANs) achieve this through an adversarial objective between a generator and discriminator~\cite{goodfellow2014gan}, while Variational Autoencoders (VAEs) generate images by decoding from a structured latent representation learned through variational inference~\cite{kingma2013autoencoding,rezende2014stochastic}. 
%Diffusion models synthesize images by iteratively  learning a sequence of conditional distributions that progressively refine noise into realistic samples~\cite{sohl2015deep,ho2020ddpm,song2021score}. Latent diffusion models (LDMs) improve efficiency by performing the diffusion process in a compressed latent space learned by an autoencoder~\cite{rombach2022ldm}. Text-conditioned generative models align image generation with semantic descriptions specified via prompts in natural-language~\cite{radford2021clip,saharia2022imagen,ramesh2022dalle}.
%\matt{Do we want to drop GAN and VAE, and begin with diffusion, then introduce VLM/VQA since that is what we use later.}\divya{Will add about VLM and VQA}
\textit{Foundation models} are large models pre-trained on massive data to learn generic representations that can be adapted to a wide range of downstream tasks with minimal supervision~\cite{bommasani2021foundation,brown2020gpt3,devlin2019bert}. \textit{Vision–Language Models (VLMs)}  learn joint representations of images and text, enabling semantic alignment between visual content and natural language descriptions~\cite{radford2021clip,jia2021align,alayrac2022flamingo,li2023blip2}. These models support multimodal reasoning and retrieval tasks such as \textit{Visual Question Answering (VQA)}, which requires them to answer natural-language questions grounded in image content~\cite{antol2015vqa,anderson2018bottomup,li2023blip2}.
Generative models learn the underlying data distribution of natural images enabling them to produce photorealistic semantic variation. \textit{Diffusion models} synthesize images by iteratively  learning a sequence of conditional distributions that progressively refine noise into realistic samples~\cite{sohl2015deep,ho2020ddpm,song2021score}. %Latent diffusion models (LDMs) improve efficiency by performing the diffusion process in a compressed latent space learned by an autoencoder~\cite{rombach2022ldm}. 
Text-conditional diffusion models~\cite{ramesh2022dalle,saharia2022imagen,nichol2021glide,radford2021clip} leverage multimodal representation to align image synthesis with semantic descriptions specified via natural-language prompts. This enables tasks such as image editing and counterfactual generation guided by natural language descriptions.

% \subsection{Semantic Knowledge Representations}
% \divya{we can cut this for space.}
% Web Ontology Language (OWL)~\cite{hogan2021knowledge} is a formal knowledge representation language used to define domain-specific structured semantics of entities and their relationships. OWL uses description logic to enable precise specification of concepts, hierarchies and constraints thereby supporting automated reasoning, consistency checking, and inference. The formal representation that captures the structure and semantics of the domain, facilitates querying, validation and extension using logical inference mechanisms.

\subsection{Semantic Feature Robustness}
Traditional notions of robustness evaluates the consistency of a model's output with respect to low-level perturbations in the input space (such as pixel-level perturbations in images)~\cite{croce2020autoattack,gowal2021improving,li2020certified,xu2025surveyadv,carlini2019evaluating,gopinath2018deepsafe}.  Robustness to more meaningful semantic features was presented most recently in ~\cite{mozumder2025rbt4dnn}. This work defines \textit{semantic-feature robustness} as the consistency of model output to semantic perturbations. 
\[
\forall x \in X,\; \forall f \in F:\;\;
N(x) = N(\Delta(x,f))
\]
where
$X$ is the input space of model $N$, 
$F$ is a set of semantic features,
and $\Delta: {X} \times {F} \rightarrow {X}$
perturbs an input with respect to a semantic feature.
While providing a general definition, the technique requires the user
to explicitly define a singleton feature model, $\lvert F \rvert = 1$, such
as the thickness of an MNIST digit.

This work also defines \textit{semantic feature functional requirement} as the consistency of model behavior with respect to a pre and postconditions. It  generates diverse inputs satisfying $\phi_x$ to check conformance of model output with respect to $\phi_y$.
\[
\forall x \in X,\; \phi_x(x) \Rightarrow \phi_y(N(x))
\]

In \textbf{\framework}, we extend this work to check the robustness of the model in satisfying the semantic feature functional requirement in the presence of semantic perturbations. 
\[
\begin{aligned}
\forall x \in X,\; \forall f \in {F}_\phi:\;\;
& (\phi_x(x) \Rightarrow \phi_y(N(x))) \Rightarrow\;\\
& (\phi_x(\Delta(x,f)) \Rightarrow \phi_y(N(\Delta(x,f))))
\end{aligned}
\]
where  ${F}_\phi \subseteq {F}$ consists of features that are irrelevant to $\phi_x$. %is the set of features orthogonal to $\Psi_x \subseteq \mathcal{F}$, the set of features relevant to the semantic property $\phi_x$;
%$\mathcal{F}_\phi \;:=\; \Psi_x^\perp$.

Our framework supports a more general, and structured, definition
of $\mathcal{F}_\phi$ which consists of an ontology of features
that can be defined by a domain-expert and expanded on during robustness
testing by foundation models.

%% file: writing/approach.tex
Figure~\ref{fig:framework} depicts the components and flow of information in \framework.
\framework accepts four primary inputs:
a neural network, $N$, under test;
a formalized correctness requirement for that network,
$\phi = (\phi_x, \phi_y)$, where when $\phi_x$ holds on inputs then $\phi_y$ must hold on corresponding outputs;
a set of inputs that satisfy the requirement, $D$ where $\forall x \in D : \phi_x(x) \Rightarrow \phi_y(N(x))$;
and a structured set of semantic features, $F_{\phi}$, under whose variation 
$\phi$ is expected to be invariant. %The elements of $F_\phi$ are initially defined by the user. 
\framework produces as output a set of pairs, $(T^i_\Delta, f^i_\Delta)$, each comprised of a
set of test inputs, $T^i_\Delta$, that violate $\phi$ and that vary from an input in $D$ with respect
to the associated feature, $f^i_\Delta$.
To generate such outputs, \framework is parameterized by a feature-perturbation operator, $\Delta : X \times F \rightarrow X$, that non-deterministically 
%\corina{what is the meaning of nondeterminism here?} \matt{I just meant that if you invoke the function with the same inputs you may get different output and we use this by generating multiple perturbations per feature/input pair} 
modifies an input with respect to a given semantic feature.

\input{writing/approach_diagram}

\framework consists of three phases that can be run repeatedly. 
The first \textit{selects features} to be perturbed within the current round, $F$, and in the first round, these are the most general features, typically defined as the roots of the hierarchical model, $F_\phi$.
The second \textit{perturbs features} $F$ present in inputs from $D$ to generate new inputs
and checks that the perturbed inputs violate $\phi$.
This yields a set, $P$ that drives the third phase. Note that we assume that $f \in F$ is orthogonal to the features impacting $\phi_x$, $\Delta(x,f)$ is a semantics-preserving transformation, therefore the perturbed input $x'$ still satisfies $\phi_x$ so $P$ is:
\begin{align*}
\{ (x, x') : \;& x' = \Delta(x,f) \wedge x \in D \wedge f \in F \wedge
\phi_x(x') \wedge \neg\phi_y(N(x'))\}
\end{align*}
The third phase analyzes the perturbed inputs in $P$ to \textit{detect features} that are common to 
groupings of failing inputs.  
This detection process is independent of $\Delta$ and can detect feature descriptions that are
not in $F_\phi$, resulting in an enriched feature space for characterizing non-robust tests.
The resulting detected features, $f^i_\Delta$, are associated with the set of inputs that were found
to be 
perturbed by that feature, $T^i_\Delta$, thereby providing a developer with actionable feedback on the causes of non-robustness in the model under test, $N$, and with witnesses to that effect.
A single pass of \framework can provide valuable and actionable information, but \framework can leverage the hierarchical structure of its feature model, $F_\phi$, to produce more focused
characterizations of the causes of non-robustness.

$F_\phi$ is organized as a set of \textit{feature trees} whose nodes represent semantic features and whose edges represent feature-refinement relations analogous to is-a and part-of relationships. Figure~\ref{fig:features} presents the feature trees for the SGSM case-study. The elements in $F_\phi$ are initially defined by the user (domain expert). Each line defines a relation among nodes in the tree -- the parent on the left and the children are operands on the right.   
A path through these relations, e.g., $\texttt{VehicleInFront}: \texttt{Type}: \texttt{Truck}$,
defines semantic information about an element in the input, e.g., that the vehicle in front is a truck, which can be used to target perturbation of the input. For a requirement $\phi$, the feature model can be defined as paths in the feature trees. The model behavior relative to $\phi$ should be invariant under changes in features at or below those paths. For instance, for the example requirement, \textit{\textbf{if a vehicle is within 10 meters, in front, and in
the same lane, the ego will not accelerate}}, shown below are a set of paths in the respective feature model.
\begin{align*}
F_\phi = \{ \;& \texttt{VehicleInFront:Type}, \\
&\texttt{VehicleInFront:Color}, \mathtt{Background} \}
\end{align*}

In subsequent rounds, the features detected as being responsible for non-robustness, $F_\Delta = \bigcup f^i_\Delta$, are used to refine $F_\phi$ and the features for the next round by 
(a) removing features for which the model was found to be robust or
% \nusrat{That is not the case always, here model was not robust for black, but it had very low failures, did not climb topK cluster and discarded. so adding the following condition}\divya{I think there was a update done here which is missing now. }
(b) deprioritizing features with low observed failure rates or
(c) defining deeper paths in the tree for features that led to non-robustness. For instance, in our example, exploring $\texttt{VehicleInFront:Color}$ yields $F_\Delta:= \{Green, Blue, Teal\}$, refining the $F_\phi$ as 
\begin{align*}
F_\phi = \{ \;& \texttt{VehicleInFront:Type}, 
\texttt{VehicleInFront:Color:Green},\\
& \texttt{VehicleInFront:Color:Blue},
\texttt{Background},\\
& \texttt{VehicleInFront:Color:Teal}\}
\end{align*}
Note that \textit{Teal} is a new feature detected by the analysis, not present in the initial tree defined by the user.

% This process terminates when either all features are removed or the leaves of the feature trees are reached.
This process terminates when all features are exhausted, the leaves of the feature trees are reached, or a user-defined bound is met.
In this way \framework produces increasingly precise explanations for non-robustness in terms of semantic features with each round. 
%\corina{might not terminate if we keep refining features: say something like: up to a user bound?}\nusrat{done}

We describe the components of \framework in more detail in the remainder of this section.

\begin{figure}[t]
\lstdefinelanguage{FeatureModel}{
  morekeywords={or,xor,not},   % add more if you wish
  sensitive=true,
  morecomment=[l]{\#},
}
\lstset{
  basicstyle=\ttfamily\small,
  keywordstyle=\bfseries\color{black},
  commentstyle=\itshape\color{gray},
  columns=fullflexible,
  showstringspaces=false,
}

\begin{lstlisting}[language=FeatureModel]
VehicleInFront = or(Type, Color, _)
VehicleToLeft = or(Type, Color, _)
Type = xor(Truck, Bus, Sedan, _)
Color = or(Black, Green, Blue, _)
Background = or(Natural, ManMade, _)
SurfaceElements = or(Natural, RoverParts, _) 
Weather = or(Dust Storm, Strong Winds, Sunglare)
\end{lstlisting}
\vspace{-3mm}
\caption{A forest of feature trees.  A feature can be defined by a mutually exclusive set of sub-features (\textbf{xor}) or as a combination of them (\textbf{or}); `\_' denotes additional unstated sub-features.\label{fig:features}}
\end{figure}

\subsection{A Forest of Feature Trees (\sfs)}

\framework processes a wide range of semantic features, where a single feature may admit a variety of perturbations and may be realized through several underlying semantic elements. 
Consequently, semantic features must be modeled at different abstraction levels, distinguishing between feature classes, which represent abstract semantic categories, and feature instances, which represent concrete realizations of those categories.  A rich body of principles and notations for describing feature models has been developed
in the Web Ontology Language (OWL). Domain experts can specify taxonomic hierarchies, compositional relations, assumptions and constraints such as cardinality, dis-jointness, or value restrictions using description logic operators~\cite{baader2003description}.  We build on these ideas
and adapt them to define a minimal language with which to express a forest of feature trees\footnote{It is a forest because the root of each tree is an independent feature.}, facilitating users to encode domain-knowledge. %\corina{should we explain why we need a "forest"?}\matt{Its only a forest because there are independent features and each has its own tree.}

\ignore{
For example, a feature class such as \textit{Vehicle Type} may be refined into specific instances such as \textit{Truck}, \textit{Bus}, or \textit{Sedan}, while another class such as \textit{Vehicle Color} may be instantiated as particular color realizations such as \textit{Yellow}, \textit{Black}, \textit{Green}, or \textit{Blue}. These feature classes represent orthogonal semantic facets describing the same object—in this case, the vehicle in front of the ego vehicle. A valid scene configuration may therefore combine instances from different feature classes, such as a \textit{Yellow Truck} or a \textit{Blue Bus}. At the same time, certain feature instances may be mutually exclusive and therefore cannot co-exist in a semantically valid configuration. For instance, a vehicle cannot simultaneously be both a \textit{Truck} and a \textit{Bus}. To systematically represent feature specialization, permissible combinations across semantic facets, and disjoint constraints while enabling structured perturbation generation, \framework organizes semantic features using a \textit{Semantic Feature Structure Graph} (\sfs), as shown in Figure~\ref{fig:sfs}.
}

All features, regardless of their level of abstraction, are defined by identifiers, e.g.,
\texttt{Color}, \texttt{Sedan}, \texttt{Natural}.  A mapping from identifiers to descriptions of
their semantics is expected to have been defined external to the system and that mapping is leveraged in the perturbation process.
A feature tree is defined by a rule:\\
\centerline{id = (\textbf{xor} | \textbf{or}) `(' id (`,' id)* (`,' `\_')? `)'}\\
where |, ?, and * are regular expression operators.
The semantics of the rules is that the feature defined on the left-hand side can be
refined to a combination of features defined by the right-hand side.
More specifically, the \textbf{or} constructor means that any subset of its operands may be present, whereas
\textbf{xor} means that at most one of its operands may be present.
%\matt{Do we need exactly one?}\nusrat{its better if we keep it, some features show mutually exclusive properties}\matt{There is a difference between mutually exclusive and the exactly one (the former allows zero).  That's what I was asking about.}\nusrat{It will be mutually exclusive, not exactly one.}
These semantics enable capturing that different types of relationships may hold between feature classes.  For example, the definition of
\texttt{Type} in Figure~\ref{fig:features} requires an exclusive or (\textbf{xor}) among \texttt{Truck}, \texttt{Bus}, and \texttt{Sedan} to model an ``is-a'' relation, whereas
the definition \texttt{Background} allows both of its children to be present akin to a ``part-of'' relation. %\corina{should we say something about how these features are produced? e.g. we start with a set of high-level features given by a domain expert?}\matt{Yes we should.  Would be good to connect to papers describing how it is plausible for domain experts to express such information, e.g., using OWL and the like.}

The `\_' character is a placeholder for an arbitrary set of unstated identifiers and is meant to support a form of \textit{open world} feature modeling.
A set of rules serves to layer depth-one trees into richer and deeper trees.
Identifiers that never appear on the right-hand side of a rule are considered \textit{roots}
of feature trees.
The full graph defined by these rules is not itself a tree, because subtrees may be shared among
roots, e.g., as in the case of \texttt{VehicleInFront} and \texttt{VehicleToLeft}.
A set of rules must be acyclic, and it follows from that that all trees are of finite depth.

As mentioned earlier, \sfs subtrees may be shared, so identifiers are not an adequate representation
of nodes in trees.  We use tree-paths to uniquely define the nodes in the trees in a forest of
feature trees.
This permits the definition of \textit{feature products},
$(\pi_1,\pi_2)$, where $\pi$ is a path from a root in the \sfs.
Products allow a form of compositional feature modeling, where instances from multiple feature classes can jointly describe the same object. For example,  the product
(\texttt{VehicleInFront}:\texttt{Type}:\texttt{Truck},\texttt{VehicleInFront}:\texttt{Color}:\texttt{Green})
states that the vehicle in front is a green truck.
Feature products correspond to logical class intersections in OWL (e.g., \textit{Type} $\cap$ \textit{Color}).

Semantic feature robustness testing becomes ill-defined if a perturbed feature is relevant to the requirement in question, since modifying such a feature may violate the requirement precondition.
For example, imagine a requirement stating that one cannot pass a stopped school bus, and the vehicle in front was perturbed to be a Sedan -- in this case, the precondition would no longer be applicable.
Since the semantics of features are described through natural language, we leave it to domain
experts to assess how those semantics interact with requirements when defining
$F_\phi$ and we assume that the roots of the defined feature trees (and subsequently the nodes appearing in the respective sub-trees) do not appear in the requirement under consideration.

\ignore{
Therefore, \framework restricts perturbations to semantic features identified by domain experts as irrelevant to the requirement. Furthermore, all descendant features of a selected node must also be irrelevant to the requirement precondition. 
Accordingly, for a requirement predicate $\phi = (\phi_x, \phi_y)$, where $\phi_x$ and $\phi_y$ denote the precondition and postcondition of the requirement respectively, the \sfs $F_{\phi}$ is defined as:
\begin{align*}
F_{\phi} \mid (\forall f \in F_{\phi}: f \in \mathrm{Irrelevant}(\phi_x))\\\
\land (\forall f' \in \mathrm{children}(f): f' \in \mathrm{Irrelevant}(\phi_x)).
\end{align*}

This formulation enables \framework to systematically generate semantically valid perturbations while ensuring that robustness analysis targets only irrelevant semantic variations. By combining ontology-based structure with explicit semantic constraints, the \sfs provides a principled mechanism for representing semantic feature relationships while maintaining perturbation feasibility and logical consistency.
}

\subsection{Perturbation and Violation Detection}
\label{ssec:pcv}
\ignore{
Given a requirement $\phi = (\phi_x, \phi_y)$, where $\phi_x$ denotes the precondition and $\phi_y$ denotes the postcondition that the model under test must satisfy, and a precondition-aware dataset $D_{\phi_x}$, \framework proceeds to evaluate the robustness of the model with respect to semantic feature perturbations. The goal of this stage is to determine whether modifying a specific semantic feature can cause the model to violate the postcondition $\phi_y$.

From the dataset $D_{\phi_x}$, we further retain only those inputs for which the model under test $N$ satisfies the postcondition predicate $\phi_y$. Formally, we construct a set $D'_{\phi_x} \subseteq D_{\phi_x}$ such that for every image $x \in D$, the prediction $N(x)$ satisfies $\phi_y$. This step ensures that any violation observed after perturbation can be attributed to the applied feature modification rather than an existing model failure. If the model already violates $\phi_y$ on an input image, further perturbations would not provide meaningful information about the influence of the targeted semantic feature.
}

\input{writing/Algo_PCV}

% \nusrat{changed the description for $P_\Delta$ here and in the algorithm. Since we use pairs from the parent features to analyze for the feature, $P_\Delta$ needs to include the ancestor pairs.}
% \divya{Algorithm 1 does not use the ancestor pairs except for a union at the end. I do not think this needs to be an input to the algorithm code since it does not impact its logic. You can mention in text that after algorithm 1 is executed, the perturbed pairs for the parent/ancestor features are also passed on to alg 2.} \corina{I agree with Divya; no need to add $P'_\Delta$ as input to teh algorithm; just say in words that you take the union; also suggest to  use $P^{anc}_\Delta$ instead of $P'_\Delta$}
Algorithm~\ref{alg:pcv} describes the procedure used to generate feature-specific perturbations and detect postcondition violations. 
The algorithm takes as input the model under test $N$, an input dataset $D$, 
a set of features $F$, 
% a feature perturbation operator $\Delta$, a postcondition predicate $\phi_y$, and the number of perturbations to generate per input $J$, 
a feature perturbation operator $\Delta$, a postcondition predicate $\phi_y$ and the number of perturbations per input $J$. 
% and a map of perturbed violating pairs over ancestor features $P'_\Delta$ that is necessary for failure-inducing feature localization\nusrat{newly added}. $P'_\Delta$ is indexed by ancestor features $g \in \mathrm{Anc}(F)$, where $\mathrm{Anc}(F) = \bigcup_{f \in F} \mathrm{Anc}(f)$ denotes all ancestors of features in $F$. $P'_\Delta[g]$ is a set of pairs, $P$, as defined earlier in Section~\ref{sec:approach}. If no ancestor exists, $P'_\Delta\gets \emptyset$\nusrat{newly added}.
% where $\mathrm{Anc}(f)$ denotes the ancestors of $f$ and $P'_\Delta = \bigcup_{f \in F}\ \bigcup_{g \in \mathrm{Anc}(f)} P_\Delta(g)$.
The algorithm produces a map, $P_\Delta$
that contains violating pairs aggregated over features 
% and their ancestor features
that is indexed by features, $f \in F $, 
where %the image of the map, 
$P_\Delta(f)$ is a set of pairs, $P$, as defined earlier in section~\ref{sec:approach}. 
%$(x,x')$, such that $x \in D$,$x'$ differs from $x$  with respect to $f$, and where $\neg \phi(x')$ whereas $\phi(x)$. 
The algorithm establishes feature-input pairs through the nested loops on
lines 2 and 4.  For each such pair, it generates $J$ perturbations by applying
the $\Delta$ operator and then records perturbations that violate the property postcondition on line 8.

For the above algorithm to be effective, $\Delta$ must be able to:
(a) produce perturbed inputs that differ from the given input only in the target feature,
(b) span a range of possible variations of that feature, and
(c) produce perturbed inputs that are realistic with respect to the dataset.
\ignore{
For each image $x \in D$, \framework generates $J$ perturbed variants by applying the perturbation operator $\mathcal{D}$ conditioned on the target feature $f$. The operator produces a perturbed image
\[
x' \leftarrow \mathcal{D}(\text{prompt}(f), x)
\]
where $\text{prompt}(f)$ specifies the desired semantic modification. The perturbed image $x'$ is then evaluated using the model $N$ to obtain the prediction $y' = N(x')$. If the prediction violates the postcondition predicate $\phi_y$, the perturbed image is added to the violated test set $T$, and the pair $(x,x')$ is inserted into the mismatch pair set $P$. This process is repeated for all perturbations of every image in $D$, after which the algorithm returns the final sets $T$ and $P$.
}
To meet these requirements, 
\framework leverages state-of-the-art diffusion models that perform instruction-guided image editing and generation. Accordingly, we define the perturbation operator as a function that invokes an editing model with the input image and a prompt that specifies the feature to be modified:
\begin{align*}
\Delta(f,x) = EditModel(x, prompt(f))
\end{align*}
where $prompt(f)$ denotes a 
% structured 
natural-language specification that encodes the intended semantic modification corresponding to feature $f$.
%recent advances in instruction-guided image editing to  that has produced AI models capable of precise semantic manipulation while preserving the remaining image content. Examples include Gemini~\cite{team2023gemini}, Flux-Kontext~\cite{labs2025flux1kontextflowmatching}, Qwen-Image-Edit-2511~\cite{wu2025qwenimagetechnicalreport}, and FLUX.2-klein~\cite{flux2-klein-9b}. 
\ignore{
In our implementation, we adopt Qwen-Image-Edit-2511 as the perturbation operator $\mathcal{D}$, as it ranked among the top-3 performing opern sourced models on the HuggingFace Artificial Analysis Image Editing Leaderboard~\cite{artificialanalysis_leaderboards} at the time of our experiments.
}
%Such editing models operate using natural language instructions that describe the desired modification. 
To ensure consistent and controlled perturbations, \framework employs a structured prompt template parameterized by natural language descriptions of feature semantics. The template constrains the editing model to modify only the target feature while preserving the remaining scene content, thereby approximating a semantics-preserving transformation. Structured prompting has been shown to improve controllability and faithfulness of generative models~\cite{brooks2023instructpix2pix,rombach2022ldm}. Please refer \cite{sefar_artifact} for more details on our prompting strategy and example prompts.
%and define a parametric-prompting approach to achieve targeted perturbation, while \corina{I cut this as it is repeated}
We rely on the generative fidelity and stochasticity of the $EditModel$ to achieve feature variation while retaining perceptual realism.

\input{writing/Alg_identify_feature}
\input{writing/requirement_tab}

\subsection{Localization of Failure-inducing Features}
\label{ssec:fet_detect}

\ignore{
The set of image pairs, $P_f$ (as shown in line 5 of Algorithm~\ref{alg:filter-features})
provide valuable evidence for identifying semantic attributes whose modification causes the model to fail. 
\framework performs semantic-feature-centric fault-localization by analyzing contrasts between inputs in $P_f$,
to determine which features are most strongly associated with observed post-condition violations. 
It also includes pairs from the ancestors of $f$, since
one or more variations of $f$ may already appear in 
those pairs\nusrat{new line}.
For example, pertubation for \texttt{VehicleInFront} can produce instances like \texttt{Blue} under \texttt{VehicleColor}.\nusrat{new line}
\divya{shouldnt the part about $P_{f}$ be further down in this section (after the below para)?}
}

Algorithm~\ref{alg:filter-features} 
uses $P_\Delta$ and $F_\phi$ to detect the set of features that best describe the causes of non-robustness.
To characterize these changes, the algorithm employs a vision-language model (VLM) $\mathcal{V}$, which maps images and text into a shared semantic embedding space and enables comparison of visual differences in terms of natural-language concepts. We use a structured prompt template to query the VLM to improve the  reliability and consistency of its outputs. Notably, we explicitly query for differences with respect to the feature $f$. Please refer~\cite{sefar_artifact} for more details.  

The algorithm generates a set of image pairs, $P_f$, 
for a feature, $f \in dom(P_\Delta)$,
by selecting from
the pairs from Algorithm~\ref{alg:pcv} those that
are associated with an
ancestors of that feature.
Including the ancestor pairs provides the opportunity to 
produce a refined analysis of the feature change.

The VLM, $\mathcal{V}$, is first used to compare $x$ and $x'$ with respect to user-defined descendants of $f$ in $F_\phi$; we use
a structured prompting approach tuned to this task.  The algorithm can handle arbitrary descendants of a feature, defined as paths through $F_\phi$ rooted at that feature.
In practice however, we have found that considering paths of length 1, $(f:c_f) \in F_\phi$, leads to reduced ambiguity in the VLM responses. When $F_\phi$ is open (i.e., contains a placeholder such as `\_' indicating unspecified features), the comparison is extended by querying the broader semantic knowledge encoded in the VLM embedding space. This enables discovery of new semantic attributes that explain differences between $x$ and $x'$. 
%\divya{Given that we only consider immediate children of $f$, is there a possibility that the discovered semantic attributes correspond to features already specified by the user but at lower levels in the tree? Will this create duplicate nodes?}
%\nusrat{It is possible to have duplicate nodes, even possible to be given by the domain expert. In this case, a feature will have multiple paths, such as \texttt{Natural}:\texttt{Rock}, \texttt{Natural}:\texttt{Mountain}:\texttt{Rock}, \texttt{Natural}:\texttt{Forest}:\texttt{Rock}}
Lines 7-10 in algorithm~\ref{alg:filter-features} capture the above mentioned logic.

%\divya{what is the necessity of $\pi$ in line 6? should it be $\pi(f)$? Is this to prevent finding fault-inducing attributes that may be higher up in the path, potentially create a cycle?}\nusrat{$\pi$ is the prompt template, I added it in the algorithm input and in the text.}\divya{as per the notations defined earlier $\pi$ is for path,and $prompt$ is the prompt function.} \nusrat{thanks for clearing, I updated} 
%\divya{I do not think prompt is an input to alg 2. It can be used inline as in the equation for the perturbation operator. Also line 9 (and possibly 8) need f to be passed to the prompt function.} \nusrat{added $f$ as prompt parameter. we can remove prompt from input if we want, I left it for others to decide.}

Once the textual descriptions of the differences between each pair %corresponding to a given feature $f$ 
are collected in $S$, \framework attempts to identify the semantic attributes that are most common (frequently occurring) in the descriptions across all pairs for that feature, $f$. Because VLMs may express the same semantic concept using different linguistic forms (such as vehicle color, color of the vehicle, or car color), direct string matching is insufficient for reliable aggregation. To address this, \framework maps each textual description into a semantic embedding space by leveraging a sentence-transformer model to capture semantic similarity between feature phrases. 
Clustering of these embeddings identifies groups of descriptions for the same underlying semantic feature. 
For instance, for the requirement and its perturbations in Section~\ref{sec:approach}, the VLM identifies \texttt{Blue} as a perturbation of \texttt{Color} when it appears in $x'$ but not in $x$, describing it as Blue, blue, or Dark Blue. \framework maps these variations into a semantic embedding space and clusters them into a single group, achieving a coverage of 69\% across image pairs.

The top-$k$ clusters (based on coverage; \# of member pairs) are then selected as the most likely explanations for differences with respect to $f$. For each cluster, the subset of perturbed inputs whose associated feature descriptions lie within the cluster is collected as the set of test cases witnessing non-robustness of $N$ with respect to variation in the corresponding semantic feature. Lines 12-16 in algorithm~\ref{alg:filter-features} capture the above mentioned logic.

\subsection{Implementation}
% \matt{We should have a short section on implementation here.}
% \nusrat{I explained details on implementation under Section~\ref{sec:evaluation}, we can summarize the writing and add here to reduce the writing size.}
% \matt{It is better to put the information related to implementation above and later here in one place.  This doesn't have to be long, but having it in one place is better.}

We select models for perturbation, feature detection, and semantic aggregation based on their effectiveness in controlled semantic manipulation and reliable extraction of feature-level changes.
For feature perturbation, we use Qwen-Image-Edit-2511~\cite{wu2025qwenimagetechnicalreport} as the operator $\Delta$, as it ranks among the top three open-source image editing models on the HuggingFace leaderboard~\cite{artificialanalysis_leaderboards}, and it outperformed alternatives like FluxKontext in preliminary studies.  For detecting semantic changes between image pairs, \framework employs the vision--language model Qwen2.5-VL-7B-Instruct~\cite{qwen2.5-VL}. To aggregate textual descriptions, we use \texttt{all-mpnet-base-v2}~\cite{reimers-2019-sentence-bert} as the sentence transformer, which encodes each description into a $768$-dimensional embedding space. 
% followed by agglomerative clustering as  $\mathcal{C}$ in Algorithm~\ref{alg:filter-features} to group semantically similar descriptions.
Furthermore, we found agglomerative clustering~\cite{sklearn_agglomerative} 
to be an effective choice for operator $\mathcal{C}$ in algorithm~\ref{alg:filter-features}. 
Additionally, we retain all non-selected test images to ensure that potential faults associated with features beyond the top-$K$ are not lost.

Both the perturbation operator and the VLM take prompts as input along with images. 
% For perturbation, we use the structured prompt template described in Section~\ref{ssec:pcv}; specifically, 
For perturbation,
we provide the prompt template and feature-specific information to ChatGPT~\cite{chatgpt}, which generates both the prompt and a corresponding negative prompt specifying undesired changes. For the VLM, we use a structured prompt template defining the comparison task, rules, and output format 
and provide it to ChatGPT to generate the final prompt for
% to ensure 
consistent extraction of semantic changes. Details of the prompts template and their generation are provided in \cite{sefar_artifact}.

%% file: writing/approach_diagram.tex
\begin{figure}[t]
\centering
\begin{tikzpicture}[
    node distance=2.2cm,
    box/.style={draw, minimum width=1cm, minimum height=1cm, align=center},
    dashedbox/.style={draw, dashed, thick, inner sep=0.3cm},
    >=latex
]

\node[box, align=center] (perb) {Perturb Features\\ and Check};
\node[box, right=0.8cm of perb, align=center] (extract) {Detect Features};
\node[box, below=0.5cm of perb, align=center] (select) {Select Features};

\node[left=0.5cm of perb, yshift=0.4cm] (lc) {$N$};
\node[left=0.5cm of perb, yshift=0.0cm] (phi) {$\phi$};
\node[left=0.5cm of perb, yshift=-0.4cm] (D) {$D$};

%\node[below of = perb, yshift=1cm] (f) {$\oplus$};

\node[left=0.5cm of select, yshift=-0.0cm] (F) {$F_{\phi}$};

\node[above of = perb, xshift=-0.0cm, yshift=-1cm] (ldm) {$\Delta$};

\node[right = 0.4cm of extract, yshift = 0mm,align=center] (out) {$(T^1_{\Delta}, f^1_{\Delta})$ \\ \ldots \\ $(T^k_{\Delta}, f^k_{\Delta})$};

\draw[->] (lc) -- (perb.west |- lc);
\draw[->] (phi) -- (perb.west |- phi);
\draw[->] (D) -- (perb.west |- D);
\draw[->] (select.north) -- node[right]{$F$} (perb.south);
\draw[->] (F) -- (select.west);
\draw[->] (ldm) -- (perb.north -| ldm);
\draw[->] (perb.east) --  node[above]{$P$} (extract.west);
\draw[->] ($(extract.east) + (0,0.)$) -- (out.west);
\draw[->] ($(out.south)+(2mm,0)$) |- node[above, xshift=-2.5cm]{$F_\Delta = \bigcup f^i_\Delta$} (select.east);

\end{tikzpicture}
\vspace{-3mm}
\caption{\framework Overview
}
\label{fig:framework}
\end{figure}

%% file: writing/Algo_PCV.tex
\begin{algorithm}[t]
\caption{\textsc{PerturbAndCheck}}
\label{alg:pcv}
\begin{algorithmic}[1]
\Require  model under test $N$; set of inputs $D$; feature set $F$, 
% Perturbed violating pairs over  ancestors of features in $F$ $P'_\Delta$ \nusrat{newly added},
feature perturbation operator $\Delta$; postcondition predicate $\phi_y$, number of perturbations $J$
\Ensure Perturbed violating pairs per feature $P_\Delta$%; violated test set $T$

% \State $P_\Delta \gets \emptyset$ %\Comment{Pairs $(x,x')$ where postcondition flips}
%\State $T \gets \emptyset$ \Comment{Perturbed tests that violate postcondition}
\State $P_\Delta \gets \emptyset$
\ForAll{$f \in F$}
    \State $P \gets \emptyset$
    \ForAll{$x \in D$} \Comment{Generate perturbations for an input-feature pair}
        \For{$j\gets 1 \text{ to } J$} 
            \State $x' \gets \Delta(f, x)$ \Comment{Generate a perturbation}
%            \State $y'_j \gets $
            \If{$\neg \phi_y(N(x'))$} \Comment{Check post-condition violation}
                \State $P \gets P \cup \{(x,x')\}$
%                \State $T \gets T \cup \{x'_j\}$
            \EndIf
        \EndFor
    \EndFor
    \State $P_\Delta[f] \gets P$ \Comment{Record pairs per feature}
\EndFor
% \State $P_\Delta \gets P_\Delta \cup P'_\Delta$ \nusrat{newly added}
\State \Return $P_\Delta$

\end{algorithmic}
\end{algorithm}

\begin{comment}
    \begin{algorithm}[t]
\caption{\textsc{Perturb} \& \textsc{Check PCV}}
\label{alg:pcv}
\begin{algorithmic}[1]
\Require  model under test $N$; set of $N$-satisfied  images $D'_{\phi_x}$; feature $f$, feature perturbation operator $\Delta$; postcondition predicate $\phi_y$,  number of perturbations $J$
\Ensure Mismatch pair set $P$; violated test set $T$

\State $P \gets \emptyset$ \Comment{Pairs $(x,x')$ where postcondition flips}
\State $T \gets \emptyset$ \Comment{Perturbed tests that violate postcondition}

\ForAll{$x \in D$}
    \For{$j\gets 1 \text{ to } J$}
        \State $x'_j \gets \Delta(prompt(f), x)$
        \State $y'_j \gets N(x'_j)$
        \If{$\neg \phi_y(y'_j)$}
            \State $P \gets P \cup \{(x,x'_j)\}$
            \State $T \gets T \cup \{x'_j\}$
        \EndIf
    \EndFor
\EndFor
\State \Return $(P, T)$

\end{algorithmic}
\end{algorithm}
\end{comment}

%% file: writing/Alg_identify_feature.tex
\begin{algorithm}[t]
\caption{\textsc{DetectViolatingFeatures}}
\label{alg:filter-features}
\begin{algorithmic}[1]
\Require Pair set $P_\Delta$; \sfs $F_{\phi}$; vision--language model $\mathcal{V}$; clustering algorithm $\mathcal{C}$; number of features $k$
\Ensure Output test-feature pairs $(T,f)$

\State $TF \gets \emptyset$

\ForAll{$f \in dom(P_\Delta)$}
    \State $S \gets \emptyset$ 
    \State $C_f \gets \{ c_f : (f:\ldots:c_f) \in F_{\phi} \}$ \Comment{Descendents of $f$}
    \State $P_f \gets \{P_\Delta[p_f] : (p_f: \ldots : f) \in F_\phi\}$\Comment{Ancestors pairs of $f$}
    \ForAll{$(x, x') \in P_f$}
        \State $S_{\text{exist}} \gets \mathcal{V}(x, x', prompt(C_f))$ \Comment{Changes to descendents}
        \State $S_{\text{new}} \gets \mathcal{V}(x, x', prompt(f))$ \Comment{All changes}
        \State $change[x'] \gets S_{\text{new}} \cup S_{\text{exist}}$ \Comment{Record changes}
        \State $S \gets S \cup change[x']$
    \EndFor

\State $\mathcal{K} \gets \mathcal{C}(S)$ \Comment{Cluster feature descriptions}
\ForAll{$c \in \mathrm{TopK}(\mathcal{K}, k)$}  \Comment{Pair tests with changes in cluster}
    \State $TF = TF \cup (\{ x' : change[x'] \subseteq c \}, \mathit{feat}(c))$ 
\EndFor \Comment{with cluster feature}

\EndFor

\State \Return $TF$
\end{algorithmic}
\end{algorithm}

% \begin{algorithm}[t]
% \caption{\textsc{ExtractSubFeatures}}
% \label{alg:rank-subfeatures}
% \begin{algorithmic}[1]
% \Require Pair set $P$ and test set $T$ for feature $f$; semantic feature tree $F_{\phi}$; coverage metric $Cov$; $k$; iteration index $i$; total iterations $I_{\max}$
% \Ensure Output feature set $F_{\text{out}}$
% % \State $D_f \gets \{x' \mid (x,x') \in P\}$ \Comment{Perturbed dataset for $f$}
% \State $C \gets \mathrm{children}_F(f)$ \Comment{Underlying subfeatures of $f$ in $F$}
% \State $S \gets \emptyset$ \Comment{List of (subfeature, coverage) pairs}
% \ForAll{$g \in C$}
%     \State $s_g \gets \mathsf{Cov}(g, T)$
%     \State $S \gets S \cup \{(g, s_g)\}$
% \EndFor
% \State $F_{\text{out}} \gets \mathrm{TopK}(S, k)$ \Comment{Top-$k$ subfeatures by coverage}
% \If{$F_{\text{out}} = \emptyset$ \textbf{and} $i < I_{\max}$}
%     \State $F_{\text{out}} \gets \textsc{FindSubFeatures}(P, f, k)$
% \EndIf
% \State \Return $F_{\text{out}}$
% \end{algorithmic}
% \end{algorithm}

%% file: writing/requirement_tab.tex
\begin{table*}
% \footnotesize
\centering
\resizebox{\textwidth}{!}{
\begin{tabular}{c|c|p{0.55\textwidth}|p{0.25\textwidth}}
     & Id & Precondition & Postcondition \\ \toprule
\multirow{4}{*}{SGSM}  & S1 & A \textcolor{red}{vehicle is within 10 meters}, \textcolor{orange}{in front}, and \textcolor{blue}{in the same lane}	& not accelerate\\ 
     & S3 & The \textcolor{red}{ego lane is controlled by a green light},  and no \textcolor{orange}{vehicle is in front}, \textcolor{blue}{in the same lane}, and \textcolor{violet}{within 10 meters}	& accelerate\\ 
     & S5 & The \textcolor{red}{ego is in the leftmost lane} and not \textcolor{orange}{in a intersection}	& not steer to the left\\ 
     & S6 & A \textcolor{red}{vehicle is in the lane to the left} and \textcolor{orange}{within 7 meters}& not steer to the left\\ 
     \midrule
 \multirow{2}{*}{RRAV}  & R1 & There is \textcolor{red}{a red or orange traffic cone in the path in front of rover} & label as traffic cone\\
     & R2 & There is a \textcolor{red}{pedestrian in the path in front of rover} 
     	& label as a pedestrian\\\midrule

AI4MARS  & A1 & The \textcolor{red}{MARS terrain contains big rocks}
& label as big rock\\
     \bottomrule
\end{tabular}}
\caption{Requirement pre and postconditions for the datasets, with semantic features highlighted in the preconditions.}
\label{tab:requirements}
\end{table*}

%% file: writing/evaluation.tex
We conduct an evaluation of \framework on models trained on three datasets, 
where each model has between one and seven requirements, and an
associated \sfs. To assess the quality and effectiveness of the generated test data in identifying feature-level robustness, we address the following research questions.

\noindent \nameref{sssec:rq1}\\
\noindent \nameref{sssec:rq2}\\
% \noindent \nameref{sssec:rq3}\\

\begin{figure*}[t]
\centering
\includegraphics[width=0.8\textwidth, height = 0.25\textwidth]{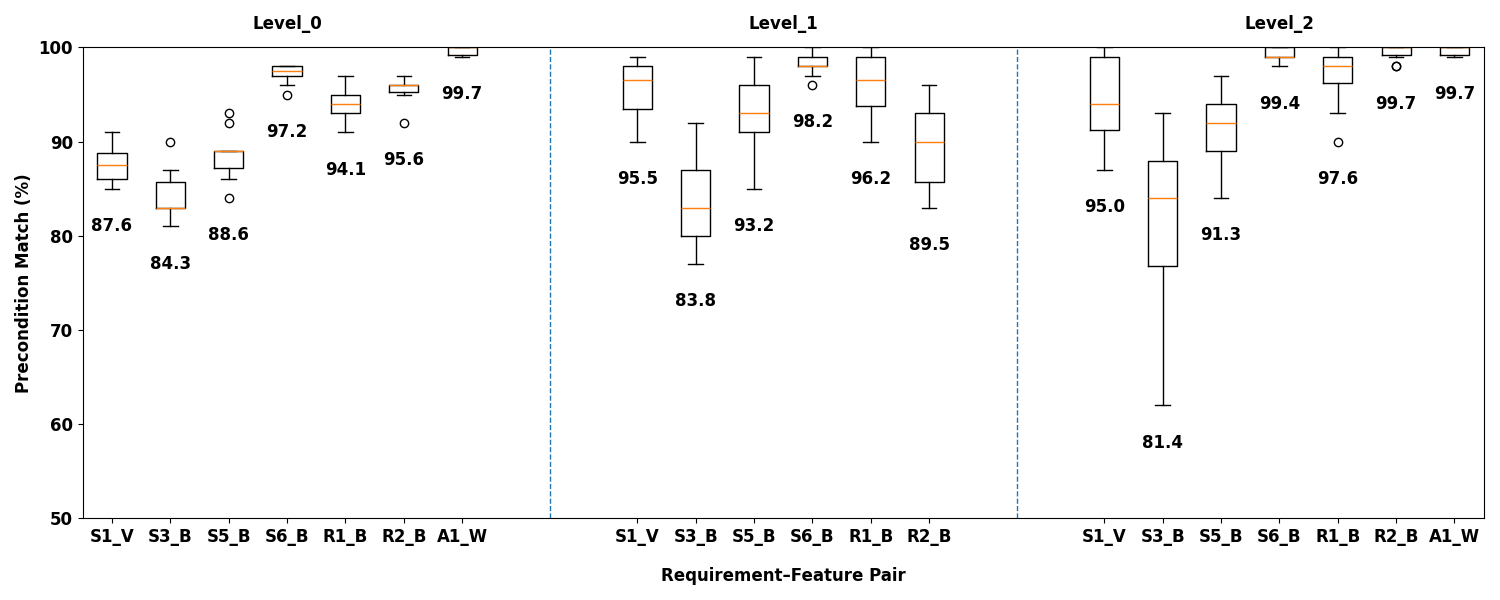}
\vspace{-3mm}
\caption{Percentage of generated tests that match the precondition. X-axis shows requirement–feature tree pairs, where each label combines a requirement (e.g., S1, S3, A1) with a feature tree: V =\texttt{VehicleInFront}, B = \texttt{Background}, and W = \texttt{Weather.}
% \matt{We should compress the vertical dimension of this plot, while keeping the fonts the same size.  It should be half the height of Fig. 7.  Needs to be done outside of latex.}
}
\label{fig:rq1}
\end{figure*}

\subsection{Evaluation Design}
\subsubsection{Dataset and Requirement Selection}

\framework can be applied to diverse settings; in this work, we evaluate it on three datasets, each representing distinct operational environments and challenges.

SGSM is an autonomous driving dataset with 10{,}885 images of resolution $900 \times 256$, captured from a forward-facing camera on a simulated ego vehicle in Town05 of the CARLA Autonomous Driving Leaderboard~\cite{carlaleaderboard}. The dataset provides scene graph abstractions encoding relationships among semantic entities, enabling structured feature reasoning. We adopt requirement specifications from~\cite{mozumder2025rbt4dnn}, derived from a subset of the Virginia Driving Code, \S46.2-842, as shown in Table~\ref{tab:requirements} (subset shown; full set in~\cite{sefar_artifact}). 

RRAV~\cite{rover} is an experimental rover developed at NASA Ames. The dataset we use for evaluation consists of images of resolution $480 \times 360$, captured from a camera mounted on the rover operating on roads and paved pathways within a campus. Due to its sequential collection process, the dataset exhibits limited diversity, with many images representing consecutive frames and appearing highly similar. 

AI4MARS~\cite{swan2021ai4mars} consists of real-world Mars terrain images collected by NASA rovers, with a resolution of $1024 \times 1024$ grayscale. The dataset includes annotations for surface types such as sand, rocks, bedrock, and big rocks, which are critical for safe navigation. %To evaluate \framework, we formulate a safety requirement targeting the detection of large rocks, stating that if the terrain contains big rocks, the model should correctly identify them (requirement A1 in Table~\ref{tab:requirements}). 

For RRAV and AI4MARS, with input from rover developers, we identified safety requirements for rovers related to obstacle avoidance.
In the case of RRAV, obstacles include traffic cones and pedestrians, while for AI4MARS, obstacles are big rocks.
These requirements are shown in Table~\ref{tab:requirements}.

To support requirement-driven analysis, we ensure sufficient data for each dataset and requirement pair, leveraging existing data analysis and generation methods when possible. For SGSM, we adopt requirement-specific fine-tuned generative models from prior work~\cite{mozumder2025rbt4dnn} as data generators. For RRAV, due to limited diversity and few samples satisfying preconditions, we follow the same approach to construct requirement-specific generators. In contrast, AI4MARS provides sufficient data, allowing us to use ground-truth annotations to filter inputs satisfying the precondition. This ensures consistent and fair evaluation of \framework across datasets with varying characteristics and data availability.

\subsubsection{Choice of \sfs Trees}

To perform targeted robustness analysis, \framework operates on selected subtrees of the \sfs (Figure~\ref{fig:features}) for each requirement, capturing semantic variations. 
The selection is guided by two criteria: (1) independence from the requirement precondition, and (2) feasibility with respect to the dataset. Independence ensures that none of the child features in the subtree influence the requirement precondition. For example, the subtree rooted at \texttt{VehicleInFront} satisfies this criterion for all SGSM and RRAV requirements, as its child features (e.g., vehicle type and color) do not affect the preconditions. Feasibility further requires that the semantic conditions needed for applying perturbations are present in the dataset; thus, this subtree is applicable only to requirement S1, which states the presence of a vehicle in front of the ego vehicle. 

Applying these criteria across datasets and requirements, we select \texttt{background} for all requirements in SGSM and RRAV, and \texttt{VehicleInFront} only for SGSM S1. For the AI4MARS dataset, we consider two subtrees in the \sfs: \texttt{Weather} and \texttt{SurfaceElements}, representing feasible semantic variations on Mars terrain that do not affect the requirement precondition. 
While these choices are illustrative of the types of feature trees a domain expert might develop, they represent a limited convenience sample of feature trees selected to facilitate this evaluation.

\subsubsection{Selection of Models to Test}
For SGSM, we adopt the model under test from prior work~\cite{mozumder2025rbt4dnn}, which uses a ResNet34-based architecture trained on the CARLA simulator data to predict continuous control signals, namely acceleration and steering angle. 

For RRAV, the model under test is a YOLOv8 model, which was trained on around 11K images of people and cones sourced from the COCO~\cite{coco2017dataset} and traffic cone TraCon~\cite{katsamenis2022tracon} datasets, with remaining data collected by operating the rover (in tele-operated mode and autonomously) on campus.  

For AI4MARS, we train a segmentation model using DeepLabV3 with the training pipeline provided in~\cite{marssimnav}. The model is trained to identify regions on terrain such as sand, rocks. We then analyse the segmentation mask to identify the segment for \texttt{Big Rock}. 
%This model achieves a validation accuracy of 95.2\%. \corina{what is the accuracy of the other models?}

% \begin{figure*}[t]
% \centering
% \includegraphics[width=0.8\textwidth, height = 0.3\textwidth]{images/rq1.png}
% \caption{Percentage of generated tests that match the precondition. X-axis shows requirement–feature tree pairs, where each label combines a requirement (e.g., S1, S3, A1) with a feature tree: V =\texttt{VehicleInFront}, B = \texttt{Background}, and W = \texttt{Weather.}
% % \matt{We should compress the vertical dimension of this plot, while keeping the fonts the same size.  It should be half the height of Fig. 7.  Needs to be done outside of latex.}
% }
% \label{fig:rq1}
% \end{figure*}

\begin{figure*}[t]
\centering
\begin{subfigure}{\linewidth}
\centering
\includegraphics[width=\linewidth]{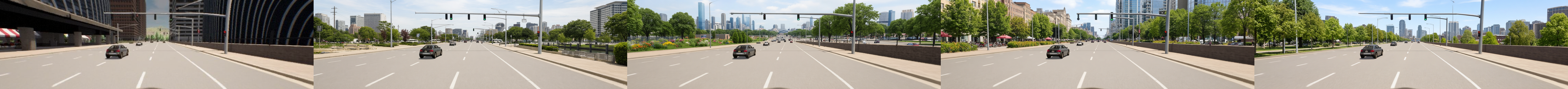}
% \caption{vehicle color change to blue}
\end{subfigure}

\begin{subfigure}{\linewidth}
\centering
\includegraphics[width=0.75\linewidth]%[width=\linewidth, height = 0.09\textwidth]
{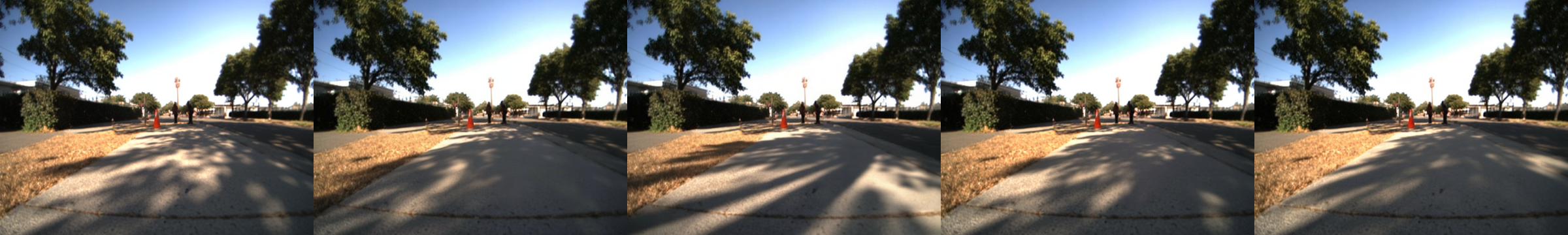}
% \caption{vehicle type change to truck}
\end{subfigure}

\begin{subfigure}{\linewidth}
\centering
\includegraphics[width=0.75\linewidth]%[width=0.85\linewidth, height = 0.1\textwidth]
{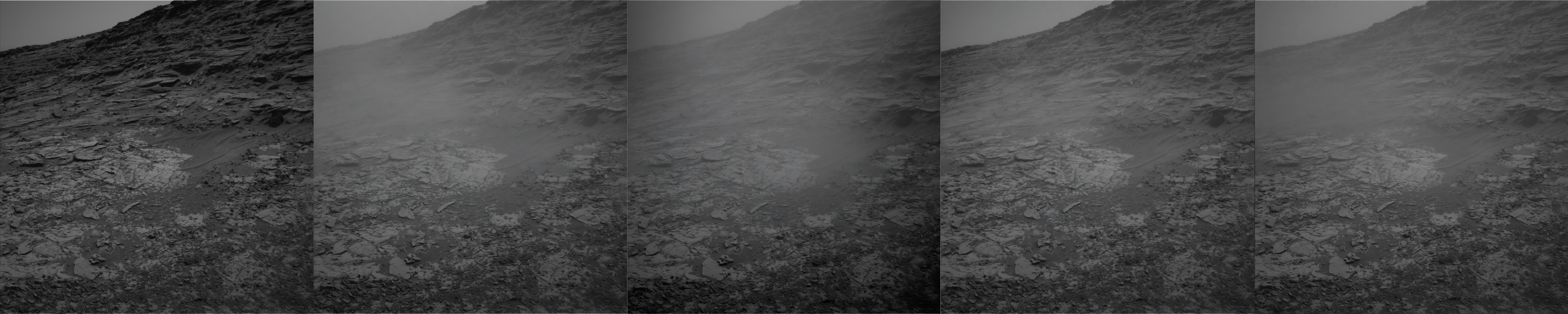}
% \caption{weather condition change to dust-storm.}
\end{subfigure}

\caption{Image panel of precondition Satisfaction. S3\_B: \texttt{Background} (top), R1\_B: \texttt{TreeShadow} (middle), A1\_W: \texttt{DustStorm} (bottom). In each row,
the leftmost image is the seed, followed by perturbed images 
that satisfy precondition, but fail the tests. 
}
\label{fig:pre_passed}

\end{figure*}

\begin{figure*}[t]
\centering
\includegraphics[width=\linewidth]%[width=0.9\textwidth, height = 0.15\textwidth]
{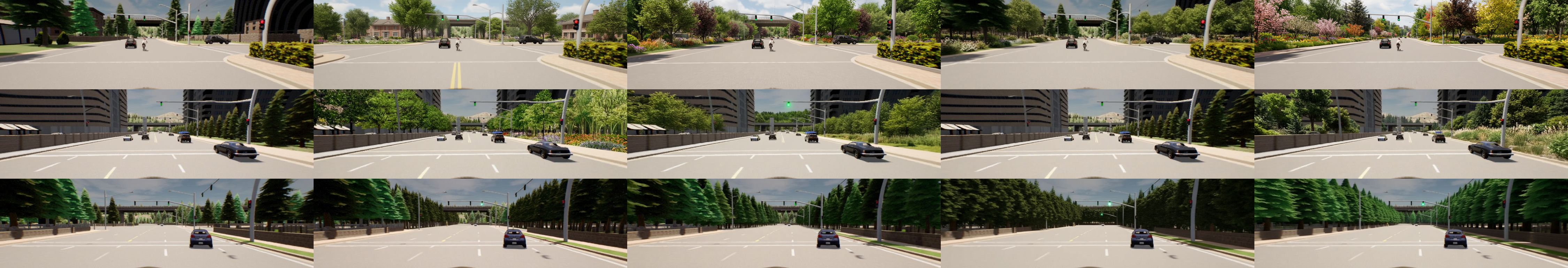}
\caption{Image panel for S3\_B: \texttt{Background} (top), \texttt{Natural} (middle), \texttt{TreeDensity} (bottom). In each row, the leftmost image is the seed, followed by perturbed images that fail the precondition classifier. 
% and also fail the postcondition check
}
\label{fig:s3_pre_violation}
\end{figure*}

\subsection{Results}
To evaluate our approach, for each requirement we start with 100 images satisfying both pre- and postconditions, perturb each image 10 times per feature, and generate 1000 distinct test cases differing in that feature; the same test set is used for all research questions. Across \sfs-requirement pairs, no root-level postcondition violations are observed for \texttt{Background} (S1, S2, S4, S7) or for \texttt{SurfaceElements} and \texttt{Weather} under A1. 
%However, visual inspection shows that \texttt{Weather} perturbations predominantly produce strong winds despite specifying its children, whereas other feature refinements exhibit variation across children. 
We exclude these cases, retain all remaining combinations (e.g., \texttt{Background} with S3, S5, S6, R1, R2), and, for \texttt{Weather}, focus perturbation on child features.

To analyze how feature granularity (i.e., global vs. specific features) affects generation and downstream analysis, we group features by levels: Level 0 (root: \texttt{Background}, \texttt{VehicleInFront}, \texttt{Weather}), Level 1 (intermediate: \texttt{Natural}, \texttt{ManMade}, \texttt{Color}, \texttt{Type}), and Level 2 (leaf: \texttt{Blue}, \texttt{Green}, and discovered features; note that \texttt{Weather} has no intermediate nodes and thus only Level 0 and Level 2). As the VLM captures fine-grained changes, unconstrained Level 0 exploration skips intermediate structure.
To preserve semantic context (e.g., whether \texttt{Natural}, \texttt{ManMade}, or both change within \texttt{Background}), we restrict Level 0 to domain-expert-defined children and explore new features within levels below.
%Level 1. Although this process can reveal deeper levels, we limit exploration to Level 2.

\begin{figure*}[t]
\centering
\includegraphics[width=0.8\textwidth, height = 0.3\textwidth]{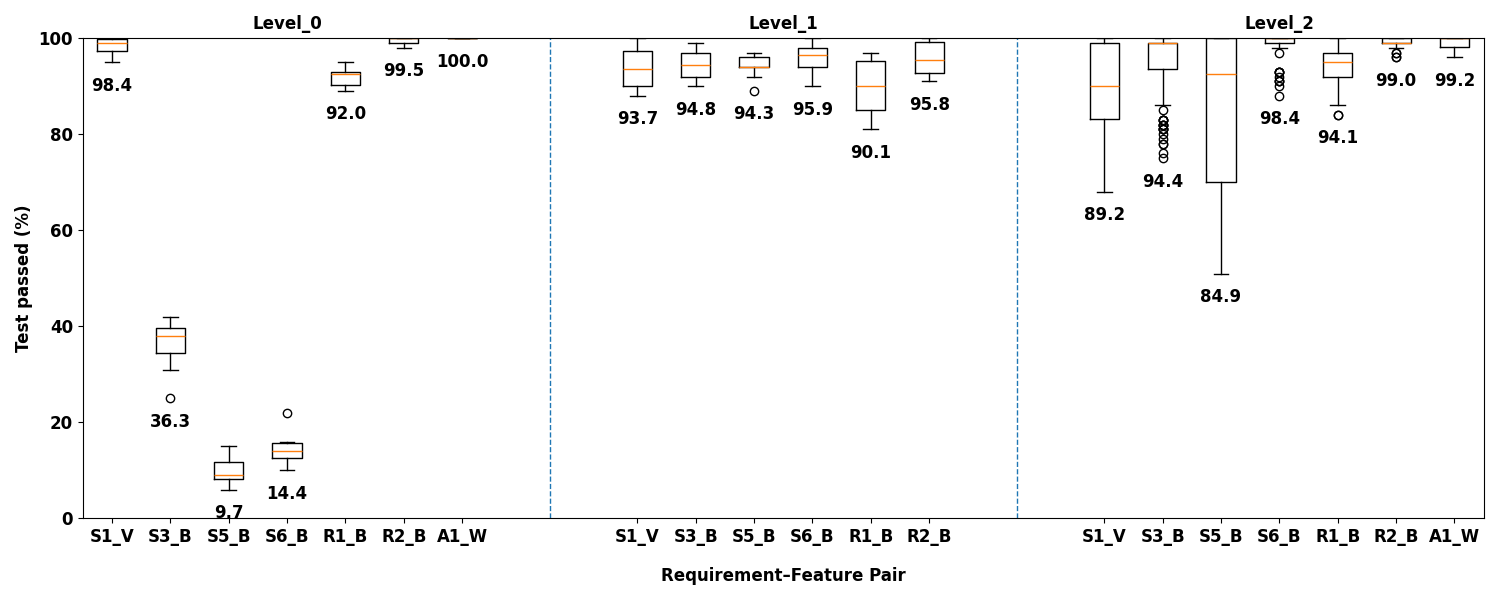}
\vspace{-3mm}
\caption{Percentage of generated tests for which the model passed. X-axis shows requirement–feature tree pairs, where each label combines a requirement (e.g., S1, S3, A1) with a feature tree: V = \texttt{VehicleInFront}, B = \texttt{Background}, and W = \texttt{Weather.}}
\label{fig:rq2}
\end{figure*}

\ignore{
\begin{figure*}[t]
\centering
\includegraphics[width=0.9\textwidth, height = 0.15\textwidth]{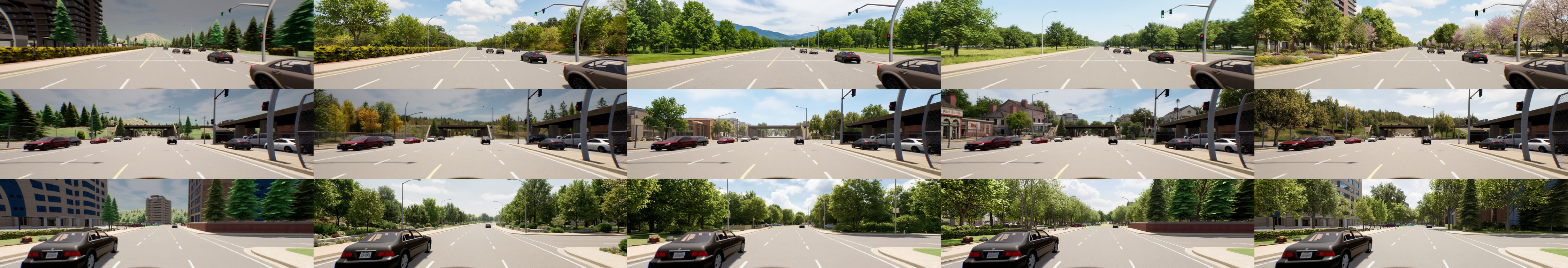}
\caption{Image panel of failed tests for feature \texttt{Background} with requirement S3 (top), S5 (middle), and S6 (bottom). In each row, the leftmost image is the seed, followed by perturbed images that failed the test.}
\label{fig:s3_s5_s6_failed}
\end{figure*}
}

\subsubsection{\rq{1}: How effective is \framework in perturbing semantic features while preserving precondition satisfaction?}
\label{sssec:rq1}

To answer this question, we assess whether \framework-generated images preserve the required preconditions while perturbing the target feature. We employ dataset-specific detection mechanisms tailored to each setting. For SGSM, we use binary classifiers from~\cite{mozumder2025rbt4dnn}, trained per requirement precondition feature to detect their presence in images (average accuracy: 94.5\%). For RRAV, we train similar classifiers using the same architecture, treating each precondition as an independent feature (average accuracy: 96.7\%). For AI4MARS, we instead use the VQA model MiniCPM-o-2\_6~\cite{MiniCPM2025} because of its strong capability of detecting objects in an image based on a question. Given the prompt “Does the image have <precondition>? Answer only yes or no,” we interpret “yes” as satisfying the precondition. The resulting precondition matches are shown in Figure~\ref{fig:rq1}.

Across all datasets, requirement-feature pairs exhibit high precondition satisfaction, with an average of 93.4\%. RRAV achieves 95.5\% and AI4MARS reaches 99.7\%, while SGSM shows the lowest average at 91.3\%. Figure~\ref{fig:pre_passed} presents example images satisfying the preconditions.

The lowest satisfaction is observed for S3\_B (S3, Background). To investigate this case, we manually analyzed samples from features with the lowest match at each level: Level 0 (\texttt{Background}: 84.3\%), Level 1 (\texttt{Natural}: 80.5\%), and Level 2 (\texttt{TreeDensity}: 71.4\%). Despite lower reported scores, visual inspection, e.g., Figure~\ref{fig:s3_pre_violation}, showed that generated images preserve the precondition, even when certain precondition features (e.g., the ego lane controlled by a green light) are not clearly visible in the seed image. The road structure and surrounding objects being unchanged suggest that the classifiers underestimate precondition satisfaction.

\ignore{
Across datasets, the trend across feature levels is not uniform but shows a consistent pattern for the challenging cases. For S3\_B, the precondition match decreases from Level 0 (84.3\%) to Level 1 (83.8\%) and further to Level 2 (81.4\%), where features become more specific (e.g., \texttt{TreeDensity}). \nusrat{should we analyze random 30 samples and report here?}Our manual inspection, however, shows that these images still preserve the precondition, suggesting that the drop is not due to actual violations but limitations of the classifiers under localized changes. In contrast, other requirement-feature pairs (e.g., S6\_B, R1\_B, R2\_B, and A1\_W) maintain consistently high satisfaction rates (above 95\%) across all levels. 
}
Across levels in the \sfs, 
the data do not reveal a clear trend in precondition satisfaction.  For some requirement-feature combinations (R2\_B), the precondition match rate increases with level, and for others (S3\_B) it decreases.
This indicates that while localized perturbations may introduce subtle variations that affect preconditions, the effect is modest, and \framework generally preserves preconditions reliably across both global and fine-grained feature manipulations.

\noindent\fbox{
    \parbox{0.46\textwidth}{
       \textbf{\rq{1} Finding: Across all datasets, requirements, and feature tree levels, \framework consistently achieves high precondition satisfaction (93.4\% on average), demonstrating its ability to perturb semantic features while preserving preconditions.}
}}

\subsubsection{\rq{2}: How effective is \framework in revealing semantic features for which models are not robust?}
\label{sssec:rq2}

To answer this question, we evaluate the model under test on the generated images and report the rate at which tests pass (\%) as judged by the postcondition in Figure~\ref{fig:rq2}. 
% For S1\_V (S1-\texttt{VehicleInFront}), the passing rate decreases consistently from Level 0 to Level 2, with the lowest performance at Level 2, with an average of 89.2\% passing rate of Level 2.
For S1\_V (S1-\texttt{VehicleInFront}), the passing rate decreases consistently from Level 0 to Level 2, reaching its lowest at Level 2, with an average passing rate of 89.2\%.
The highest failure rates are observed for vehicle color \texttt{Green} (39.7\%), \texttt{Blue} (11.3\%), and vehicle type \texttt{Truck} (10.5\%), showing that \framework effectively identifies individual features that significantly influence model decisions.
Combining features (e.g., \texttt{GreenTruck}) results in a 25.5\% failure rate. 
For \texttt{Weather}, only \texttt{DustStorm} results in failures (2.4\%), while \texttt{StrongWinds} and \texttt{Sunglare} show no failures, consistent with the 100\% passing rate for \texttt{Weather} in Figure~\ref{fig:rq2}, where most images correspond to \texttt{StrongWinds}.

For \texttt{Background}, particularly under S3, S5, and S6, Figure~\ref{fig:rq2} shows a sharp drop in passing rate at Level 0 
(e.g., S3\_B: 36.3\%, S5\_B: 9.7\%, S6\_B: 14.4\%). Such low performance can arise from either (1) false positives due to unmet preconditions or (2) a lack of model robustness. However, Figure~\ref{fig:rq1} reports high precondition match rates (around 90\%) for these cases, and manual inspection (S3\_B of Figure~\ref{fig:pre_passed}) confirms that the generated images satisfy preconditions, eliminating the first possibility.

Examining deeper levels in Figure~\ref{fig:rq2}, the passing rate improves significantly for intermediate nodes (\texttt{Natural}, \texttt{ManMade}) and remains relatively high for most leaf features, indicating that not all feature changes equally impact the model. However, certain discovered features show notable failures: \texttt{Lighting} (under \texttt{Natural}) causes 19.2\% and 29.1\% failures for S3 and S5, respectively, while \texttt{TrafficLightPole} (under \texttt{ManMade}) leads to 26.7\% failures for S5. When such features are perturbed jointly, the failure rates increase substantially: for S5, \texttt{Lighting}+\texttt{TrafficLightPole} results in 51.3\% failures, and \texttt{SkyLight}+\texttt{Building} results in 40.8\%, indicating that combined semantic changes amplify model vulnerability.

A similar trend is observed for RRAV, where intermediate level perturbations reveal more failures than individual leaf features, indicating that multi-feature semantic changes have a stronger impact on model robustness. This further demonstrates that \framework effectively identifies and isolates influential feature interactions.

\noindent\fbox{
    \parbox{0.46\textwidth}{
       \textbf{\rq{2} Finding: \framework generated tests effectively reveal model sensitivity to individual and composite features, with an average of 16.3\% of the tests revealing feature non-robustness.}
}} 

%% file: writing/threats.tex
While our evaluation spans multiple datasets, models, and requirements, like all experiments, it is limited by pragmatic choices that may impact conclusions.
Wherever possible, we leveraged neural networks, datasets, and requirements from other
sources and from domain experts to mitigate bias in our study.

We selected features to model for each requirement that we felt were \textit{non-controversial}, e.g., that the presence of a big rock should be independent of weather, that a driving maneuver should be independent of vehicle color.
A broader study of \framework leveraging domain ontologies written by experts is
a natural next step to broaden the strength of the findings of this work.

\framework's ability to detect non-robustness and diagnose features that cause it depends on the quality of foundation models and prompting strategies.
We experimented with different models and strategies, but we fully expect
that models will improve in their ability to produce realistic images and target
semantics-preserving changes based on text prompts.  These will only improve the effectiveness of
\sfs by further increasing precondition match rates and targeting a broader
range of features.

%% file: writing/Conclusion.tex
\framework proposes a novel approach to systematically evaluate model robustness to requirement-independent semantic features by generating controlled, precondition-preserving perturbations and iteratively refining feature hierarchies to uncover fault-inducing descendants. Our results show that it reliably produces valid test cases while precisely isolating features that expose model vulnerabilities, enabling effective and fine-grained semantic robustness analysis. These findings highlight the importance of semantic feature-level testing and position \framework as a practical tool for improving the reliability of vision-based systems.
%\noindent\textbf{Acknowledgment.} We thank NASA’s RRAV team for providing the experimental rover data and model.

%% file: writing/Data_availability.tex
The implementation of \framework and relative data can be found here: https://doi.org/10.5281/zenodo.19341589.

%% file: main.bib
@article{mozumder2025rbt4dnn,
  title={RBT4DNN: Requirements-based Testing of Neural Networks},
  author={Mozumder, Nusrat Jahan and Toledo, Felipe and Dola, Swaroopa and Dwyer, Matthew B},
  journal={arXiv preprint arXiv:2504.02737},
  year={2025}
}

@inproceedings{woodlief2024s3c,
  title={S3c: Spatial semantic scene coverage for autonomous vehicles},
  author={Woodlief, Trey and Toledo, Felipe and Elbaum, Sebastian and Dwyer, Matthew B},
  booktitle={Proceedings of the IEEE/ACM 46th International Conference on Software Engineering},
  pages={1--13},
  year={2024}
}

@inproceedings{attaoui2025designator,
  author       = {Mohammed Oualid Attaoui and Fabrizio Pastore},
  title        = {DESIGNATOR: a Toolset for Automated GAN-enhanced Search-based Testing and Retraining of DNNs in Martian Environments},
  booktitle    = {Proceedings of the 40th IEEE/ACM International Conference on Automated Software Engineering (ASE 2025) Tool Demonstration Track},
  year         = {2025},
}

@article{AttaouiPB25,
  author       = {Mohammed Oualid Attaoui and
                  Fabrizio Pastore and
                  Lionel C. Briand},
  title        = {Search-Based {DNN} Testing and Retraining With GAN-Enhanced Simulations},
  journal      = {{IEEE} Trans. Software Eng.},
  volume       = {51},
  number       = {4},
  pages        = {1086--1103},
  year         = {2025},
}

@inproceedings{baresi2025efficient,
  author       = {Luciano Baresi and Davide Yi Xian Hu and Andrea Stocco and Paolo Tonella},
  title        = {Efficient Domain Augmentation for Autonomous Driving Testing Using Diffusion Models},
  booktitle    = {Proceedings of the 47th International Conference on Software Engineering (ICSE 2025)},
  year         = {2025},
}

@inproceedings{duenkel2025cns,
  author    = {D{\"u}nkel, Olaf and Jesslen, Artur and Xie, Jiahao and Theobalt, Christian and Rupprecht, Christian and Kortylewski, Adam},
  title     = {CNS-Bench: Benchmarking Image Classifier Robustness Under Continuous Nuisance Shifts},
  booktitle = {Proceedings of the IEEE/CVF International Conference on Computer Vision (ICCV)},
  year      = {2025},
}

@inproceedings{Mofayezi_2023_CVPR,
  author    = {Mofayezi, Mohammadreza and Medghalchi, Yasamin},
  title     = {Benchmarking Robustness to Text-Guided Corruptions},
  booktitle = {Proceedings of the IEEE/CVF Conference on Computer Vision and Pattern Recognition (CVPR) Workshops},
  year      = {2023},
}

@inproceedings{ConceptAlgebra2023,
  author       = {Yilun Du and
                  Shuang Li and
                  Antonio Torralba and
                  Joshua B. Tenenbaum},
  title        = {Concept Algebra for Fine-Grained Control in Diffusion Models},
  booktitle    = {Advances in Neural Information Processing Systems (NeurIPS)},
  year         = {2023}
}

@inproceedings{Conceptor2024,
  author       = {Daniel Rueckert and
                  Michael Niemeyer and
                  Andrea Vedaldi and
                  Christian Rupprecht},
  title        = {Conceptor: Concept-Oriented Controllable Generation for Diffusion Models},
  booktitle    = {International Conference on Learning Representations (ICLR)},
  year         = {2024}
}

@inproceedings{SeFa2021,
  author       = {Yujun Shen and
                  Ceyuan Yang and
                  Xiaoou Tang and
                  Bolei Zhou},
  title        = {Closed-Form Factorization of Latent Semantics in GANs},
  booktitle    = {IEEE Conference on Computer Vision and Pattern Recognition (CVPR)},
  pages        = {1532--1540},
  year         = {2021}
}

@inproceedings{NoiseCLR2024,
  author       = {Melanie Subedar and
                  Karl Ridgeway and
                  Andrew Gordon Wilson and
                  Han Liu and
                  Jun-Yan Zhu},
  title        = {NoiseCLR: Noise-Guided Disentangled Representation Learning for Diffusion Models},
  booktitle    = {IEEE/CVF Conference on Computer Vision and Pattern Recognition (CVPR)},
  year         = {2024}
}

@inproceedings{Geodesics2025,
  author       = {Firstname Lastname and
                  Firstname Lastname and
                  Firstname Lastname},
  title        = {Riemannian Geodesics in Diffusion Generative Models},
  booktitle    = {IEEE/CVF Conference on Computer Vision and Pattern Recognition (CVPR)},
  year         = {2025}
}

@inproceedings{prabhu2023lance,
  title     = {LANCE: Stress-testing Visual Models by Generating Language-guided Counterfactual Images},
  author    = {Viraj Prabhu and Sriram Yenamandra and Prithvijit Chattopadhyay and Judy Hoffman},
  booktitle = {Advances in Neural Information Processing Systems 36 (NeurIPS 2023)},
  year      = {2023}
}

@inproceedings{Zhang_2024_ImageNetD,
  author    = {Zhang, Chenshuang and Pan, Fei and Kim, Junmo and Kweon, In So and Mao, Chengzhi},
  title     = {ImageNet-D: Benchmarking Neural Network Robustness on Diffusion Synthetic Object},
  booktitle = {Proceedings of the IEEE/CVF Conference on Computer Vision and Pattern Recognition (CVPR)},
  year      = {2024}
}

@article{scenic-mlj23,
  author    = {Daniel J. Fremont and Edward Kim and Tommaso Dreossi and
               Shromona Ghosh and Xiangyu Yue and Alberto L. Sangiovanni-Vincentelli
               and Sanjit A. Seshia},
  title     = {Scenic: A Language for Scenario Specification and Data Generation},
  journal   = {Machine Learning},
  volume    = {112},
  number    = {10},
  pages     = {3805--3849},
  year      = {2023},
  publisher = {Springer}
}

@inproceedings{verifai-cav19,
  author    = {Tommaso Dreossi and Daniel J. Fremont and Shromona Ghosh and
               Edward Kim and Hadi Ravanbakhsh and Marcell Vazquez-Chanlatte and
               Sanjit A. Seshia},
  title     = {{VerifAI}: A Toolkit for the Formal Design and Analysis of
               Artificial Intelligence-Based Systems},
  booktitle = {Proceedings of the 31st International Conference on Computer
               Aided Verification (CAV)},
  year      = {2019},
  organization = {Springer},
  series    = {Lecture Notes in Computer Science},
  volume    = {11561},
  pages     = {343--353}
}

@inproceedings{HeRGT23,
  author    = {Yejun He and Muslim Razi and Jerry Zeyu Gao and Chuanqi Tao},
  title     = {A Framework for Autonomous Vehicle Testing Using Semantic Models},
  booktitle = {Proceedings of the IEEE International Conference on Artificial Intelligence Testing (AITest)},
  year      = {2023},
  pages     = {66--73},
  doi       = {10.1109/AITest58265.2023.00020}
}

@article{AlnaserSA21,
  author    = {Ala' J. Alnaser and Arman Sargolzaei and Mustafa Ilhan Akbas and Rishi Razdan and Rodney Sell and Mark Bellone and Michael Menase and Mehran Malayjerdi},
  title     = {Autonomous Vehicles Scenario Testing Framework and Model of Computation: On Generation and Coverage},
  journal   = {IEEE Access},
  volume    = {9},
  pages     = {60617--60628},
  year      = {2021},
  doi       = {10.1109/ACCESS.2021.3074062}
}

@inproceedings{FremontKPS20,
  author    = {Daniel J. Fremont and Edward Kim and Yash Vardhan Pant and Sanjit A. Seshia and Atul Acharya and Xantha Bruso and Paul Wells and Steve Lemke and Qiang Lu and Shalin Mehta},
  title     = {Formal Scenario-Based Testing of Autonomous Vehicles: From Simulation to the Real World},
  booktitle = {2020 IEEE 23rd International Conference on Intelligent Transportation Systems (ITSC)},
  year      = {2020},
  pages     = {1--10},
  doi       = {10.1109/ITSC45102.2020.9294368}
}

@inproceedings{goodfellow2015explaining,
  title     = {Explaining and Harnessing Adversarial Examples},
  author    = {Goodfellow, Ian J. and Shlens, Jonathon and Szegedy, Christian},
  booktitle = {Proceedings of the International Conference on Learning Representations (ICLR)},
  year      = {2015},
  note      = {arXiv:1412.6572}
}

@inproceedings{kurakin2017adversarial,
  title     = {Adversarial Examples in the Physical World},
  author    = {Kurakin, Alexey and Goodfellow, Ian and Bengio, Samy},
  booktitle = {Proceedings of the International Conference on Learning Representations (ICLR), Workshop Track},
  year      = {2017},
  note      = {arXiv:1607.02533}
}

@inproceedings{papernot2017practical,
  title     = {Practical Black-Box Attacks against Machine Learning},
  author    = {Papernot, Nicolas and McDaniel, Patrick and Goodfellow, Ian and Jha, Somesh and Celik, Z. Berkay and Swami, Ananthram},
  booktitle = {Proceedings of the 2017 ACM Asia Conference on Computer and Communications Security (AsiaCCS)},
  pages     = {506--519},
  year      = {2017},
  publisher = {ACM},
  doi       = {10.1145/3052973.3053009}
}

@inproceedings{hendrycks2019benchmarking,
  title     = {Benchmarking Neural Network Robustness to Common Corruptions and Perturbations},
  author    = {Hendrycks, Dan and Dietterich, Thomas},
  booktitle = {Proceedings of the International Conference on Learning Representations (ICLR)},
  year      = {2019},
  note      = {IMAGENET-C and IMAGENET-P benchmarks}
}

@inproceedings{pei2017deepxplore,
  title     = {DeepXplore: Automated Whitebox Testing of Deep Learning Systems},
  author    = {Pei, Kexin and Cao, Yinzhi and Yang, Junfeng and Jana, Suman},
  booktitle = {Proceedings of the 26th ACM Symposium on Operating Systems Principles (SOSP)},
  pages     = {1--18},
  year      = {2017},
  publisher = {ACM},
  doi       = {10.1145/3132747.3132785}
}

@misc{wu2025qwenimagetechnicalreport,
      title={Qwen-Image Technical Report}, 
      author={Chenfei Wu and Jiahao Li and Jingren Zhou and Junyang Lin and Kaiyuan Gao and Kun Yan and Sheng-ming Yin and Shuai Bai and Xiao Xu and Yilei Chen and Yuxiang Chen and Zecheng Tang and Zekai Zhang and Zhengyi Wang and An Yang and Bowen Yu and Chen Cheng and Dayiheng Liu and Deqing Li and Hang Zhang and Hao Meng and Hu Wei and Jingyuan Ni and Kai Chen and Kuan Cao and Liang Peng and Lin Qu and Minggang Wu and Peng Wang and Shuting Yu and Tingkun Wen and Wensen Feng and Xiaoxiao Xu and Yi Wang and Yichang Zhang and Yongqiang Zhu and Yujia Wu and Yuxuan Cai and Zenan Liu},
      year={2025},
      eprint={2508.02324},
      archivePrefix={arXiv},
      primaryClass={cs.CV},
      url={https://arxiv.org/abs/2508.02324}, 
}

@misc{qwen2.5-VL,
    title = {Qwen2.5-VL},
    url = {https://qwenlm.github.io/blog/qwen2.5-vl/},
    author = {Qwen Team},
    month = {January},
    year = {2025}
}

@misc{chatgpt,
  author       = {{OpenAI}},
  title        = {ChatGPT},
  year         = {2025},
  howpublished = {\url{https://chat.openai.com}},
  note         = {Accessed: 2026-03-13}
}

@misc{artificialanalysis_leaderboards,
  author       = {{Artificial Analysis}},
  title        = {Artificial Analysis AI Model Leaderboards},
  year         = {2025},
  howpublished = {\url{https://huggingface.co/spaces/ArtificialAnalysis/Text-to-Image-Leaderboard}},
  note         = {Accessed: 2026-03-13}
}

@misc{qwen_huggingface,
  author       = {{Qwen Team}},
  title        = {Qwen Models on Hugging Face},
  year         = {2025},
  howpublished = {\url{https://huggingface.co/Qwen}},
  note         = {Accessed: 2026-03-13}
}

@misc{MiniCPM2025,
  author       = "{OpenBMB}",
  title        = "{MiniCPM-o-2\_6}",
  year         = {2025},
  url          = {https://huggingface.co/openbmb/MiniCPM-o-2_6},
  note         = {Accessed: March 3, 2026}
}

@article{hu2024corruption,
  title={Assessing Visually Continuous Corruption Robustness of Neural Networks Relative to Human Performance},
  author={Hu, Boyue Caroline and Chechik, Marsha},
  journal={arXiv preprint arXiv:2402.19401},
  year={2024}
}

@inproceedings{deepxplore2017,
  title={DeepXplore: Automated Whitebox Testing of Deep Learning Systems},
  author={Pei, Kexin and others},
  booktitle={SOSP},
  year={2017}
}

@inproceedings{ma2018deepgauge,
  title={DeepGauge: Multi-granularity Testing Criteria for Deep Learning Systems},
  author={Ma, Lei and others},
  booktitle={ASE},
  year={2018}
}

@inproceedings{deeproad2018,
  title={DeepRoad: GAN-based Metamorphic Autonomous Driving System Testing},
  author={Zhang, Mengshi and others},
  booktitle={ASE},
  year={2018}
}

@misc{flux2024,
  title={FLUX: Text-to-Image Generation Model},
  author={Black Forest Labs},
  year={2024},
  note={\url{https://huggingface.co/black-forest-labs}}
}

@misc{fluxkontext2024,
  title={FLUX-Kontext: Context-aware Image Editing with Diffusion Models},
  author={Black Forest Labs},
  year={2024},
  note={\url{https://huggingface.co/black-forest-labs}}
}

@article{qwenVL2023,
  title={Qwen-VL: A Versatile Vision-Language Model for Understanding, Localization, and Generation},
  author={Bai, Jinze and others},
  journal={arXiv preprint arXiv:2308.12966},
  year={2023}
}

@article{conceptsliders2024,
  title={Concept Sliders: Interpretable Control of Diffusion Models},
  author={Gandikota, Rohit and others},
  journal={arXiv preprint arXiv:2304.xxxxx},
  year={2024}
}

@inproceedings{yin2024pascalEA,
  title={Benchmarking Semantic Segmentation Models via Appearance and Geometry Attribute Editing},
  author={Yin, et al.},
  booktitle={CVPR},
  year={2024}
}

@article{dillema2025,
  title={DILLEMA: Diffusion and Large Language Models for Multi-Modal Augmentation},
  author={Dei PoliMi authors},
  year={2025}
}

@inproceedings{conceptualedits2025,
  title={Visual Counterfactual Explanations via Conceptual Edits},
  booktitle={NeurIPS},
  year={2025}
}

@inproceedings{fathi2024decodex,
  title={DeCoDEx: Confounder Detector Guidance for Diffusion-based Counterfactual Explanations},
  author={Fathi, Nima and others},
  booktitle={Proceedings of Machine Learning Research (PMLR)},
  year={2024}
}

@misc{carlaleaderboard,
  key = {CARLA Leaderboard},
  title = {{CARLA Leaderboard 1.0 – SENSORS Track (0.9.10.1)}},
  howpublished = "\url{https://leaderboard.carla.org/leaderboard/}",
  note = "[Online; accessed 23-Oct-2024]"
}

@article{qwenImage2025,
  title={Qwen-Image Technical Report},
  author={{Qwen Team}},
  journal={arXiv preprint arXiv:2508.02324},
  year={2025},
  url={https://arxiv.org/abs/2508.02324}
}

@inproceedings{croce2020autoattack,
  title={Reliable evaluation of adversarial robustness with an ensemble of diverse attacks},
  author={Croce, Francesco and Hein, Matthias},
  booktitle={ICML},
  year={2020}
}

@inproceedings{gowal2021improving,
  title={Improving robustness using generated data},
  author={Gowal, Sven and Qin, Chongli and Uesato, Jonathan and Mann, Timothy and Kohli, Pushmeet},
  booktitle={NeurIPS},
  year={2021}
}

@inproceedings{li2020certified,
  title={Certified robustness for deep neural networks},
  author={Li, Linyi and Xie, Tao and Li, Bo},
  booktitle={IEEE Symposium on Security and Privacy},
  year={2020}
}

@article{xu2025surveyadv,
  title={A Survey of Adversarial Examples in Computer Vision},
  author={Xu, Keyizhi and others},
  journal={Wuhan University Journal of Natural Sciences},
  year={2025}
}

@article{carlini2019evaluating,
  title={On Evaluating Adversarial Robustness},
  author={Carlini, Nicholas and others},
  journal={arXiv preprint arXiv:1902.06705},
  year={2019}
}

@inproceedings{gopinath2018deepsafe,
  title={DeepSafe: A Data-Driven Approach for Assessing Robustness of Neural Networks},
  author={Gopinath, Divya and Katz, Guy and Pasareanu, Corina S. and Barrett, Clark},
  booktitle={International Symposium on Automated Technology for Verification and Analysis (ATVA)},
  pages={3--19},
  year={2018},
  publisher={Springer}
}

@inproceedings{zhang2023unsupervised,
  title={Towards Unsupervised Object Detection from LiDAR Point Clouds},
  author={Zhang, Lunjun and Yang, Anqi Joyce and Xiong, Yuwen and Casas, Sergio and Yang, Bin and Ren, Mengye and Urtasun, Raquel},
  booktitle={Proceedings of the IEEE/CVF Conference on Computer Vision and Pattern Recognition (CVPR)},
  pages={9317--9328},
  year={2023}
}

@inproceedings{wu2022trajectoryguided,
  title={Trajectory-guided Control Prediction for End-to-end Autonomous Driving: A Simple yet Strong Baseline},
  author={Wu, Penghao and Jia, Xiaosong and Chen, Li and Yan, Junchi and Li, Hongyang and Qiao, Yu},
  booktitle={Advances in Neural Information Processing Systems (NeurIPS)},
  volume={35},
  pages={6119--6132},
  year={2022}
}

@inproceedings{shao2023safety,
  title={Safety-Enhanced Autonomous Driving Using Interpretable Sensor Fusion Transformer},
  author={Shao, Huajian and Wang, Lei and Chen, Rui and Li, Hongyang and Liu, Yu},
  booktitle={Proceedings of the Conference on Robot Learning (CoRL)},
  pages={726--737},
  year={2023},
  organization={PMLR}
}

@inproceedings{toledo2021deeper,
  title={Deeper Notions of Correctness for Machine Learning-based Systems},
  author={Toledo, Fernando and Shriver, Dylan and Elbaum, Sebastian and Dwyer, Matthew B.},
  booktitle={Proceedings of the International Conference on Software Engineering (ICSE)},
  year={2021}
}

@inproceedings{bojarski2016end,
  title={End to End Learning for Self-Driving Cars},
  author={Bojarski, Mariusz and Del Testa, Davide and Dworakowski, Daniel and Firner, Bernhard and Flepp, Beat and Goyal, Prasoon and Jackel, Larry and Monfort, Mathew and Muller, Urs and Zhang, Jiakai and others},
  booktitle={Proceedings of the IEEE Conference on Computer Vision and Pattern Recognition Workshops (CVPRW)},
  year={2016}
}

@inproceedings{chen2020learning,
  title={Learning by Cheating},
  author={Chen, Dian and Koltun, Vladlen and Kr{\"a}henb{\"u}hl, Philipp},
  booktitle={Proceedings of the IEEE/CVF Conference on Computer Vision and Pattern Recognition (CVPR)},
  pages={5127--5136},
  year={2020}
}

@article{proenca2019deep,
  title={Deep Learning for Spacecraft Pose Estimation from Monocular Images},
  author={Proenca, Pedro and Gao, Yang},
  journal={IEEE Transactions on Aerospace and Electronic Systems},
  volume={56},
  number={2},
  pages={1497--1510},
  year={2019}
}

@inproceedings{van2018satellite,
  title={You Only Look Twice: Rapid Multi-Scale Object Detection In Satellite Imagery},
  author={Van Etten, Adam},
  booktitle={Proceedings of the IEEE Conference on Computer Vision and Pattern Recognition (CVPR)},
  year={2018}
}

@inproceedings{demir2018deepglobe,
  title={DeepGlobe 2018: A Challenge to Parse the Earth through Satellite Images},
  author={Demir, Ilke and Koperski, Krzysztof and Lindenbaum, David and Pang, Guan and Huang, Jianbo and Basu, Saikat and Hughes, Forest and Tuia, Devis and Raska, Rafael},
  booktitle={Proceedings of the IEEE Conference on Computer Vision and Pattern Recognition (CVPR) Workshops},
  year={2018}
}

@article{zhu2017deep,
  title={Deep Learning in Remote Sensing: A Comprehensive Review and List of Resources},
  author={Zhu, Xiao Xiang and Tuia, Devis and Mou, Lichao and Xia, Gui-Song and Zhang, Liangpei and Xu, Feng and Fraundorfer, Friedrich},
  journal={IEEE Geoscience and Remote Sensing Magazine},
  volume={5},
  number={4},
  pages={8--36},
  year={2017}
}

@inproceedings{hundman2018detecting,
  title={Detecting Spacecraft Anomalies Using LSTMs and Nonparametric Dynamic Thresholding},
  author={Hundman, Kyle and Constantinou, Valentino and Laporte, Christopher and Colwell, Ian and Soderstrom, Tom},
  booktitle={Proceedings of the 24th ACM SIGKDD International Conference on Knowledge Discovery and Data Mining (KDD)},
  year={2018}
}

@article{silva2026crash,
  title={CRASH: Cognitive Reasoning Agent for Safety Hazards in Autonomous Driving},
  author={Silva, Erick and Yasmin, Rehana and Shoker, Ali},
  journal={arXiv preprint arXiv:2603.15364},
  year={2026}
}

@article{abdelaty2024accident,
  title={A Matched Case-Control Analysis of Autonomous vs Human-driven Vehicle Accidents},
  author={Abdel-Aty, Mohamed and others},
  journal={Nature Communications Engineering},
  year={2024}
}

@article{penmetsa2021crashes,
  title={Effects of Autonomous Vehicle Crashes on Public Perception of Self-Driving Technology},
  author={Penmetsa, Pavan and others},
  journal={Transportation Research Part F},
  year={2021}
}

@article{banks2017tesla,
  title={Driver Error or Designer Error: Using the Perceptual Cycle Model to Explore the Circumstances Surrounding the Fatal Tesla Crash},
  author={Banks, Victoria and Plant, Katherine and Stanton, Neville},
  journal={Transportation Research Part F},
  year={2017}
}

@standard{iso21448,
  title        = {Road Vehicles -- Safety of the Intended Functionality},
  organization = {International Organization for Standardization},
  number       = {ISO/PAS 21448:2019},
  year         = {2019},
  note         = {SOTIF standard for safety of autonomous driving systems}
}

@inproceedings{azzalini2023eventbased,
  title     = {On the Generation of Synthetic Event-Based Vision Datasets for Navigation and Landing},
  author    = {Azzalini, Loïc J. and Blazquez, Emmanuel and Hadjiivanov, Alexander and Meoni, Gabriele and Izzo, Dario},
  booktitle = {Proceedings of the ESA Guidance, Navigation and Control Conference (ESA GNC)},
  pages     = {1--14},
  year      = {2023}
}

@inproceedings{zysk2023infrared,
  title     = {Generation of Artificial Infrared Camera Images for Visual Navigation Simulation},
  author    = {Zysk, Krystian and Hałoń, Michał and Kaczmarek, Kacper and Kasprzyk, Marcin and Rodo, Piotr and Skromak, Olgierd and Sochacki, Mateusz},
  booktitle = {Proceedings of the ESA Guidance, Navigation and Control Conference (ESA GNC)},
  address   = {Sopot, Poland},
  year      = {2023}
}

@article{zohdinasab2022feature,
  title={Efficient and Effective Feature Space Exploration for Testing Deep Learning Systems},
  author={Zohdinasab, Tina and Riccio, Vincenzo and Gambi, Alessandra and Tonella, Paolo},
  journal={ACM Transactions on Software Engineering and Methodology (TOSEM)},
  year={2022}
}

@inproceedings{wang2022bet,
  title={BET: Black-box Efficient Testing for Convolutional Neural Networks},
  author={Wang, Jianwen and Qiu, Huili and Rong, Yujia and Ye, Hongyu and Li, Qiang and Li, Zhen and Zhang, Chao},
  booktitle={Proceedings of the 31st ACM SIGSOFT International Symposium on Software Testing and Analysis (ISSTA)},
  pages={164--175},
  year={2022}
}

@inproceedings{lee2020adaptive,
  title={Effective White-Box Testing of Deep Neural Networks with Adaptive Neuron-Selection Strategy},
  author={Lee, Seokhyeon and Cha, Seongjoon and Lee, Donghwan and Oh, Heungsun},
  booktitle={Proceedings of the 29th ACM SIGSOFT International Symposium on Software Testing and Analysis (ISSTA)},
  pages={165--176},
  year={2020}
}

@inproceedings{guo2018dlfuzz,
  title={DLFuzz: Differential Fuzzing Testing of Deep Learning Systems},
  author={Guo, Jianjun and Jiang, Yuchi and Zhao, Yue and Chen, Quan and Sun, Jianjun},
  booktitle={Proceedings of the 26th ACM Joint Meeting on European Software Engineering Conference and Symposium on the Foundations of Software Engineering (ESEC/FSE)},
  pages={739--743},
  year={2018}
}

@inproceedings{tian2018deeptest,
  title={DeepTest: Automated Testing of Deep-Neural-Network-Driven Autonomous Cars},
  author={Tian, Yuchi and Pei, Kexin and Jana, Suman and Ray, Baishakhi},
  booktitle={Proceedings of the 40th International Conference on Software Engineering (ICSE)},
  pages={303--314},
  year={2018}
}

@inproceedings{dola2024cit4dnn,
  title={Cit4DNN: Generating Diverse and Rare Inputs for Neural Networks using Latent Space Combinatorial Testing},
  author={Dola, Sumanth and McDaniel, Reuben and Dwyer, Matthew B. and Soffa, Mary Lou},
  booktitle={Proceedings of the IEEE/ACM 46th International Conference on Software Engineering (ICSE)},
  pages={1--13},
  year={2024}
}

@inproceedings{swan2021ai4mars,
  title={AI4MARS: A Dataset for Terrain-Aware Autonomous Driving on Mars},
  author={Swan, R. Michael and Atha, Deegan and Leopold, Henry A. and Gildner, Matthew and Oij, Stephanie and Chiu, Cindy and Hong, Xiangyu and Ono, Masahiro},
  booktitle={Proceedings of the IEEE/CVF Conference on Computer Vision and Pattern Recognition Workshops (CVPRW)},
  year={2021}
}

@inproceedings{toledo2024sgsm,
  title={Specifying and Monitoring Safe Driving Properties with Scene Graphs},
  author={Toledo, Fernando and Woodlief, Thomas and Elbaum, Sebastian and Dwyer, Matthew B.},
  booktitle={Proceedings of the IEEE International Conference on Robotics and Automation (ICRA)},
  pages={15577--15584},
  year={2024},
  organization={IEEE}
}

@misc{marssimnav,
  author       = {{Pushkar Hue}},
  title        = {MarsSimNav: Mars Simulation Navigation Dataset and Tools},
  year         = {2025},
  howpublished = {\url{https://github.com/pushkar-hue/MarsSimNav}},
  note         = {Accessed: 2026-03-13}
}

@report{NTSB2019UberCrash,
  author = {{National Transportation Safety Board}},
  title = {Collision Between Vehicle Controlled by Developmental Automated Driving System and Pedestrian},
  institution = {NTSB},
  year = {2019},
  number = {HWY18MH010},
  note = {Uber self-driving crash, Tempe, Arizona}
}

@report{NTSB2020TeslaCrash,
  author = {{National Transportation Safety Board}},
  title = {Collision Between a Sport Utility Vehicle Operating With Partial Driving Automation and a Crash Attenuator},
  institution = {NTSB},
  year = {2020},
  number = {HWY19FH007},
  note = {Tesla Autopilot crash, Mountain View, California}
}

@report{NTSB2021TeslaTruck,
  author = {{National Transportation Safety Board}},
  title = {Collision Between a Car Operating With Automated Vehicle Control Systems and a Tractor-Semitrailer Truck},
  institution = {NTSB},
  year = {2021},
  note = {Failure to detect truck crossing highway}
}

@report{NTSB2018TeslaFire,
  author = {{National Transportation Safety Board}},
  title = {Battery Fire After High-Speed Collision Involving Electric Vehicle},
  institution = {NTSB},
  year = {2018},
  note = {Post-crash battery fire risk highlighting perception and response limitations}
}

@misc{rover,
  title = {N{A}{S}{A} Experimental rover},
  howpublished = {\url{https://ntrs.nasa.gov/api/citations/20250004071/downloads/RRAV_2025AmesParternshipDays.mp4}},
  note = {Accessed: 2026-01-20}
}

@inproceedings{giusti2016machine,
  title={Machine learning for autonomous navigation in unstructured environments},
  author={Giusti, Alessandro and Guzzi, J{\'e}r{\^o}me and Ciresan, Dan C and He, Fang-Lin and Rodriguez, Juan P and Fontana, Flavio and Faessler, Matthias and Forster, Christian and Schmidhuber, J{\"u}rgen and Di Caro, Gianni A and others},
  booktitle={IEEE International Conference on Robotics and Automation (ICRA)},
  year={2016}
}

@inproceedings{loquercio2018dronet,
  title={DroNet: Learning to fly by driving},
  author={Loquercio, Antonio and Maqueda, Ana I and del-Blanco, Carlos R and Scaramuzza, Davide},
  booktitle={IEEE Robotics and Automation Letters (RA-L)},
  year={2018}
}

@article{pham2018deep,
  title={Deep learning for real-time UAV navigation and obstacle avoidance},
  author={Pham, Huy and La, Hung},
  journal={IEEE Transactions on Intelligent Transportation Systems},
  year={2018}
}

@inproceedings{sohl2015deep,
  title={Deep Unsupervised Learning using Nonequilibrium Thermodynamics},
  author={Sohl-Dickstein, Jascha and Weiss, Eric and Maheswaranathan, Niru and Ganguli, Surya},
  booktitle={ICML},
  year={2015}
}

@inproceedings{song2021score,
  title={Score-Based Generative Modeling through Stochastic Differential Equations},
  author={Song, Yang and Sohl-Dickstein, Jascha and Kingma, Diederik and Kumar, Abhishek and Ermon, Stefano and Poole, Ben},
  booktitle={ICLR},
  year={2021}
}

@inproceedings{radford2021clip,
  title={Learning Transferable Visual Models From Natural Language Supervision},
  author={Radford, Alec and Kim, Jong Wook and Hallacy, Chris and Ramesh, Aditya and Goh, Gabriel and Agarwal, Sandhini and Sastry, Girish and Askell, Amanda and Mishkin, Pamela and Clark, Jack and others},
  booktitle={ICML},
  year={2021}
}

@inproceedings{saharia2022imagen,
  title={Photorealistic Text-to-Image Diffusion Models with Deep Language Understanding},
  author={Saharia, Chitwan and Chan, William and Saxena, Saurabh and Li, Lala and Whang, Jay and Denton, Emily and Ghasemipour, Kamyar and Ayan, Burcu Karagol and Mahdavi, Sepehr and Lopes, Raphael and others},
  booktitle={NeurIPS},
  year={2022}
}

@inproceedings{ramesh2022dalle,
  title={Hierarchical Text-Conditional Image Generation with CLIP Latents},
  author={Ramesh, Aditya and Dhariwal, Prafulla and Nichol, Alex and Chu, Casey and Chen, Mark},
  booktitle={arXiv preprint arXiv:2204.06125},
  year={2022}
}

@misc{coco2017dataset,
  title={COCO 2017 Dataset},
  author={{COCO Consortium}},
  howpublished={\url{https://cocodataset.org}},
  year={2017}
}

@article{katsamenis2022tracon,
  title={TraCon: A novel dataset for real-time traffic cones detection using deep learning},
  author={Katsamenis, Iason and Karolou, Eleni Eirini and Davradou, Agapi and Protopapadakis, Eftychios and Doulamis, Anastasios and Doulamis, Nikolaos and Kalogeras, Dimitris},
  journal={arXiv preprint arXiv:2205.11830},
  year={2022}
}

@article{bommasani2021foundation,
  title={On the Opportunities and Risks of Foundation Models},
  author={Bommasani, Rishi and Hudson, Drew A. and Adeli, Ehsan and Altman, Russ and Arora, Simran and von Arx, Sydney and Bernstein, Michael S. and Bohg, Jeannette and Bosselut, Antoine and Brunskill, Emma and others},
  journal={arXiv preprint arXiv:2108.07258},
  year={2021}
}

@article{brown2020gpt3,
  title={Language Models are Few-Shot Learners},
  author={Brown, Tom B. and Mann, Benjamin and Ryder, Nick and Subbiah, Melanie and Kaplan, Jared and Dhariwal, Prafulla and Neelakantan, Arvind and Shyam, Pranav and Sastry, Girish and Askell, Amanda and others},
  journal={Advances in Neural Information Processing Systems (NeurIPS)},
  volume={33},
  pages={1877--1901},
  year={2020}
}

@inproceedings{devlin2019bert,
  title={BERT: Pre-training of Deep Bidirectional Transformers for Language Understanding},
  author={Devlin, Jacob and Chang, Ming-Wei and Lee, Kenton and Toutanova, Kristina},
  booktitle={Proceedings of NAACL-HLT},
  pages={4171--4186},
  year={2019}
}

@inproceedings{jia2021align,
  title={Scaling Up Visual and Vision-Language Representation Learning With Noisy Text Supervision},
  author={Jia, Chao and Yang, Yinfei and Xia, Ye and Chen, Yen-Chun and Parekh, Zarana and Pham, Hieu and Le, Quoc V. and Sung, Yun-Hsuan and Li, Zhen and Duerig, Thomas},
  booktitle={Proceedings of ICML},
  year={2021}
}

@article{alayrac2022flamingo,
  title={Flamingo: a Visual Language Model for Few-Shot Learning},
  author={Alayrac, Jean-Baptiste and Donahue, Jeff and Luc, Pauline and Miech, Antoine and Barr, Iain and Hasson, Yana and Lenc, Karel and Mensch, Arthur and Millican, Katie and Reynolds, Matthew and others},
  journal={Advances in Neural Information Processing Systems (NeurIPS)},
  year={2022}
}

@article{li2023blip2,
  title={BLIP-2: Bootstrapping Language-Image Pre-training with Frozen Image Encoders and Large Language Models},
  author={Li, Junnan and Li, Dongxu and Savarese, Silvio and Hoi, Steven C. H.},
  journal={Proceedings of ICML},
  year={2023}
}

@inproceedings{antol2015vqa,
  title={VQA: Visual Question Answering},
  author={Antol, Stanislaw and Agrawal, Aishwarya and Lu, Jiasen and Mitchell, Margaret and Batra, Dhruv and Zitnick, C. Lawrence and Parikh, Devi},
  booktitle={Proceedings of ICCV},
  pages={2425--2433},
  year={2015}
}

@inproceedings{anderson2018bottomup,
  title={Bottom-Up and Top-Down Attention for Image Captioning and Visual Question Answering},
  author={Anderson, Peter and He, Xiaodong and Buehler, Chris and Teney, Damien and Johnson, Mark and Gould, Stephen and Zhang, Lei},
  booktitle={Proceedings of CVPR},
  year={2018}
}

@article{ho2020ddpm,
  title={Denoising Diffusion Probabilistic Models},
  author={Ho, Jonathan and Jain, Ajay and Abbeel, Pieter},
  journal={Advances in Neural Information Processing Systems (NeurIPS)},
  year={2020}
}

@inproceedings{rombach2022ldm,
  title={High-Resolution Image Synthesis with Latent Diffusion Models},
  author={Rombach, Robin and Blattmann, Andreas and Lorenz, Dominik and Esser, Patrick and Ommer, Bj{\"o}rn},
  booktitle={Proceedings of CVPR},
  year={2022}
}

@article{nichol2021glide,
  title={GLIDE: Towards Photorealistic Image Generation and Editing with Text-Guided Diffusion Models},
  author={Nichol, Alex Quinn and Dhariwal, Prafulla and Ramesh, Aditya and Shyam, Pranav and Mishkin, Pamela and McGrew, Bob and Sutskever, Ilya and Chen, Mark},
  journal={arXiv preprint arXiv:2112.10741},
  year={2021}
}

@book{baader2003description,
  title={The Description Logic Handbook: Theory, Implementation, and Applications},
  author={Baader, Franz and Calvanese, Diego and McGuinness, Deborah and Nardi, Daniele and Patel-Schneider, Peter},
  publisher={Cambridge University Press},
  year={2003}
}

@inproceedings{brooks2023instructpix2pix,
  title={InstructPix2Pix: Learning to Follow Image Editing Instructions},
  author={Brooks, Tim and Holynski, Aleksander and Efros, Alexei A.},
  booktitle={Proceedings of the IEEE/CVF Conference on Computer Vision and Pattern Recognition (CVPR)},
  pages={18392--18402},
  year={2023}
}

@misc{sefar_artifact,
  author       = {{Anonymous Authors}},
  title        = {{SEFAR Artifact Repository}},
  howpublished = {\url{https://doi.org/10.5281/zenodo.19341589}},
  year         = {2026},
  note         = {Accessed: 2026-03-30}
}

@inproceedings{reimers-2019-sentence-bert,
    title = "Sentence-BERT: Sentence Embeddings using Siamese BERT-Networks",
    author = "Reimers, Nils and Gurevych, Iryna",
    booktitle = "Proceedings of the 2019 Conference on Empirical Methods in Natural Language Processing",
    month = "11",
    year = "2019",
    publisher = "Association for Computational Linguistics",
    url = "https://arxiv.org/abs/1908.10084",
}

@misc{sklearn_agglomerative,
  author       = {{Scikit-learn Developers}},
  title        = {{AgglomerativeClustering — scikit-learn documentation}},
  howpublished = {\url{https://scikit-learn.org/stable/modules/generated/sklearn.cluster.AgglomerativeClustering.html}},
  year         = {2026},
  note         = {Accessed: 2026-03-25}
}
